\documentclass{article}

\usepackage[main, final]{neurips_2026}

\usepackage[utf8]{inputenc} 
\usepackage[T1]{fontenc}    
\usepackage{hyperref}       
\usepackage{url}            
\usepackage{booktabs}       
\usepackage{amsfonts}       
\usepackage{nicefrac}       
\usepackage{microtype}      
\usepackage{xcolor}         

\usepackage{wrapfig}
\usepackage{graphicx}
\usepackage{multirow}
\usepackage{enumitem}
\usepackage[table]{xcolor}
\newcommand{\model}{REEF}
\usepackage{amsmath}

\usepackage{booktabs}
\usepackage{amssymb}
\usepackage{caption}

\title{
Relation-Aware Graph Foundation Model
}

\author{%
Jianxiang Yu\textsuperscript{1},
Jiapeng Zhu\textsuperscript{1},
Yibo Zhao\textsuperscript{1},
Hao Qian\textsuperscript{2},
Ziqi Liu\textsuperscript{2},
Zhiqiang Zhang\textsuperscript{2},
Xiang Li\textsuperscript{1}
\\
\textsuperscript{1}School of Data Science and Engineering, East China Normal University \\
\textsuperscript{2}Ant Group \\
\texttt{\{jianxiangyu,jiapengzhu,yibozhao\}@stu.ecnu.edu.com} \\
\texttt{\{qianhao.qh,ziqiliu,lingyao.zzq\}@antgroup.com} \\
\texttt{xiangli@dase.ecnu.edu.cn} \\
}

\begin{document}

\maketitle

\begin{abstract}
 In recent years, large language models (LLMs) have demonstrated remarkable capability to generalize across diverse natural language processing tasks,
inspiring the development of graph foundation models (GFMs) for large-scale pre-training. 
However, 
unlike language models with explicit token units,
graphs lack a well-defined unit for generalization, making it challenging to design effective pre-training strategies. 
In this work, we propose \model, 
a novel GFM framework that leverages relation tokens as the fundamental units.
We construct a 
vocabulary 
of
relation tokens to encode relational information within graphs.
To accommodate diverse relations, 
we introduce two hypernetworks that adaptively generate the parameters of aggregators and classifiers in graph neural networks based on relation tokens. 
In addition,
we design another hypernetwork to construct dataset-specific projectors and incorporate a dataset-level feature bias into the initial node representations,
enhancing flexibility across different datasets with the same relation. 
Extensive experiments demonstrate that \model~consistently outperforms existing methods in both pre-training and transfer learning, 
highlighting its potential as a general-purpose graph foundation model.
Our code is publicly available at \url{
https://github.com/jianxiangyu/REEF/}.
\end{abstract}

\section{Introduction}
In recent years, large language models (LLMs) have gained significant attention for their powerful generalization capabilities~\citep{achiam2023gpt, wei2022emergent,touvron2023llama,brown2020language}.
These models, referred to as foundation models, are pre-trained on extensive data, allowing them to learn diverse linguistic patterns and effectively adapt to a wide range of downstream tasks.

\begin{wrapfigure}{r}{0.55\textwidth}
 \vspace{-1.0em}
  \centering
  \includegraphics[width=1.0\linewidth]{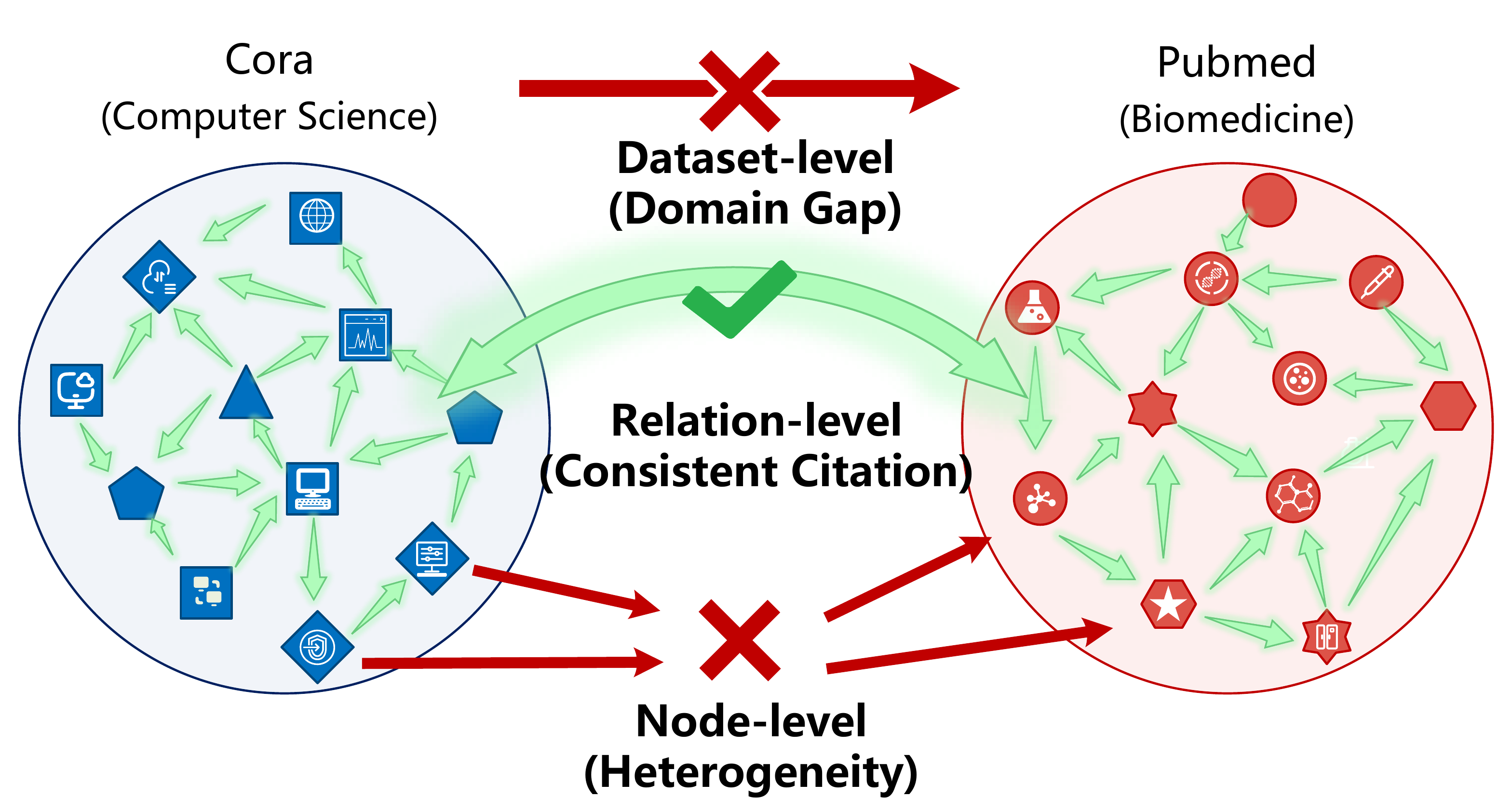}
  \caption{Comparison of transfer unit at the node, dataset, and relation levels. }
  \label{fig:unit_compare}
  \vspace{-1.0em}
\end{wrapfigure}

Inspired by these advances,
the graph learning community is exploring the development of graph foundation models (GFMs), which aim to generalize across diverse datasets through large-scale pre-training and to adapt to few-shot settings with limited supervision.
Graphs, as universal data structures, naturally capture entities and their \emph{relations},
which encode both structural and semantic interactions and form the connective backbone of graph data.
For example, a sentence can be represented as a dependency graph where relations encode grammatical roles~\citep{gao2021abcd}; an image can be modeled as a graph where pixels are connected through spatial adjacency~\citep{wu2020comprehensive}; and in real-world systems such as citation or social networks~\citep{liu2022deep,myers2014information}, 
relations define how information 
flows among entities.  
Unlike language, which explicitly decomposes text into tokens as fundamental units, graph data lacks such clear units of generalization. 
This absence presents a key challenge for GFMs, making it difficult to design effective pre-training strategies.

Existing attempts often treat either nodes or datasets as fundamental units for pre-training~\citep{huang2024prodigy,zhao2024all,liu2024one,xia2024anygraph}.
These approaches either learn distinct node embeddings or rely on mixture-of-experts (MoE) frameworks to handle different datasets.
Despite exhibiting some degree of generalization, these methods are limited in capturing universal knowledge.
As illustrated in Figure~\ref{fig:unit_compare},
citation networks like Cora and PubMed belong to different domains (computer science vs. biomedicine) and 
exhibit drastic differences at both the node and dataset levels.
In contrast,
their \emph{relations} consistently capture citation interactions between nodes across datasets.
\textbf{
Therefore, relations offer a more coherent and transferable basis for GFMs, making them better suited as fundamental units for pre-training.
}

\begin{wrapfigure}{r}{0.7\textwidth}
 \vspace{-0.8em}
  \centering
  \includegraphics[width=1.0\linewidth]{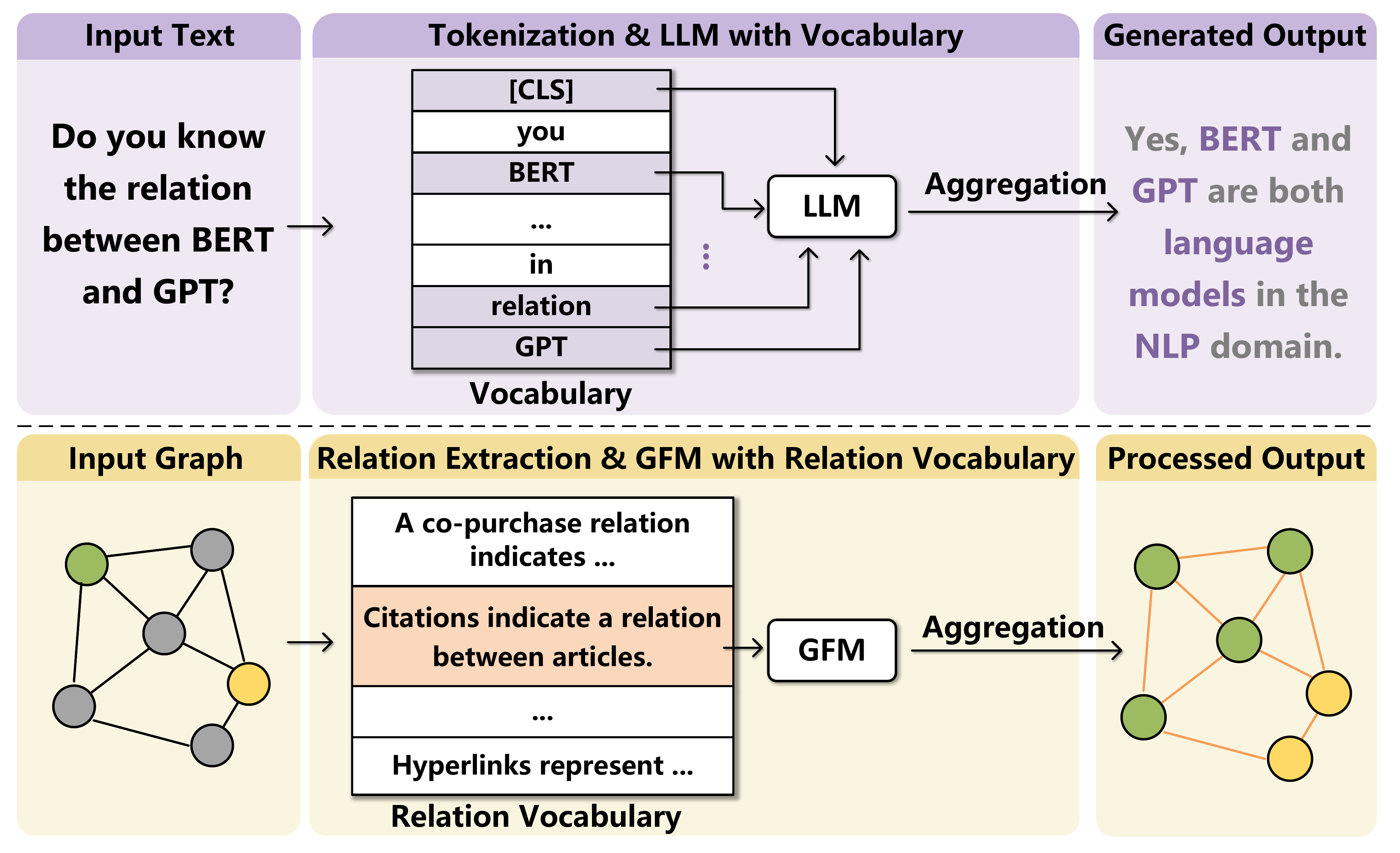}
  \caption{Relation vocabulary inspired by LLMs. }
  \label{fig:relation_vocab}
\end{wrapfigure}


Our inspiration stems from how LLMs extract linguistic information from tokens,
under the premise that a predefined vocabulary encodes each individual token.
To comprehend an input sequence and generate appropriate responses, LLMs learn to aggregate information from tokens through weighted attention, yielding a global semantic representation used for output generation.
With this idea in mind, we propose extending this paradigm to the graph domain, as depicted in Figure~\ref{fig:relation_vocab}. 
Concretely, we analogously construct a vocabulary of \emph{relation tokens} that encode relational information within graphs, serving as the fundamental units for GFMs.
For a given input graph, each type of edge is mapped to a relation token from this vocabulary, capturing the semantics of the relation. 
A global graph representation is then derived by aggregating messages from the source and target nodes of each relation, weighted by the GFM parameters learned during pretraining.
In essence, our approach focuses on modeling relations through flexible, token-specific message aggregation.

Based on this motivation,
we propose a novel framework called~\model,
which utilizes \textbf{RE}lation \textbf{E}ncoding to construct a graph \textbf{F}oundation model.
We conceptualize graph mining tasks as the modeling of diverse relations, where link prediction or classification involves determining whether a relation exists—either between two nodes or between a node and a label.
To accommodate various types of relationships, we customize key components of graph neural networks (GNNs), including the \emph{aggregator} and \emph{classifier}. 
Specifically, we first build a relation vocabulary using language models, encoding textual descriptions of relationships into relational representations. 
These representations are then used to parameterize two \emph{hypernetworks}~\citep{chauhan2024brief,ha2016hypernetworks}, which adaptively generate relation-specific aggregators and classifiers, enabling flexible message passing and effective task-specific inference.
Further,
to handle diverse feature distributions
across datasets within the same relation,
we introduce an additional hypernetwork
to construct 
dataset-specific
projectors 
and incorporate dataset-level bias into the initial node representations.
Finally, we adopt pre-training with mixed-dataset training and graph data augmentation strategies,
enabling~\model~to capture relational diversity more effectively.
Extensive experimental results demonstrate that 
\model~achieves superior performance on both pre-training and transfer learning tasks,
and delivers particularly substantial gains in few-shot transfer scenarios.
Our main contributions are summarized as follows:

\begin{itemize}[leftmargin=*, itemsep=0pt, parsep=0pt, topsep=0.2pt, partopsep=0pt]
    \item We propose a novel framework for graph foundation models that leverages relation tokens as the fundamental units, enabling effective pre-training and transfer across datasets.
    \item 
    We design two hypernetworks to dynamically generate relation-specific aggregators and classifiers based on relational representations. 
    Additionally, we introduce dataset-specific feature projectors and bias terms to enhance representation quality and improve generalization capabilities.
    \item 
    We conduct extensive experiments demonstrating that REEF achieves state-of-the-art average performance across datasets, 
    with especially strong improvements in zero-shot and few-shot transfer scenarios.
\end{itemize}
\section{Preliminary}

\begin{figure*}[t]
    \centering
\includegraphics[width=0.99\linewidth]{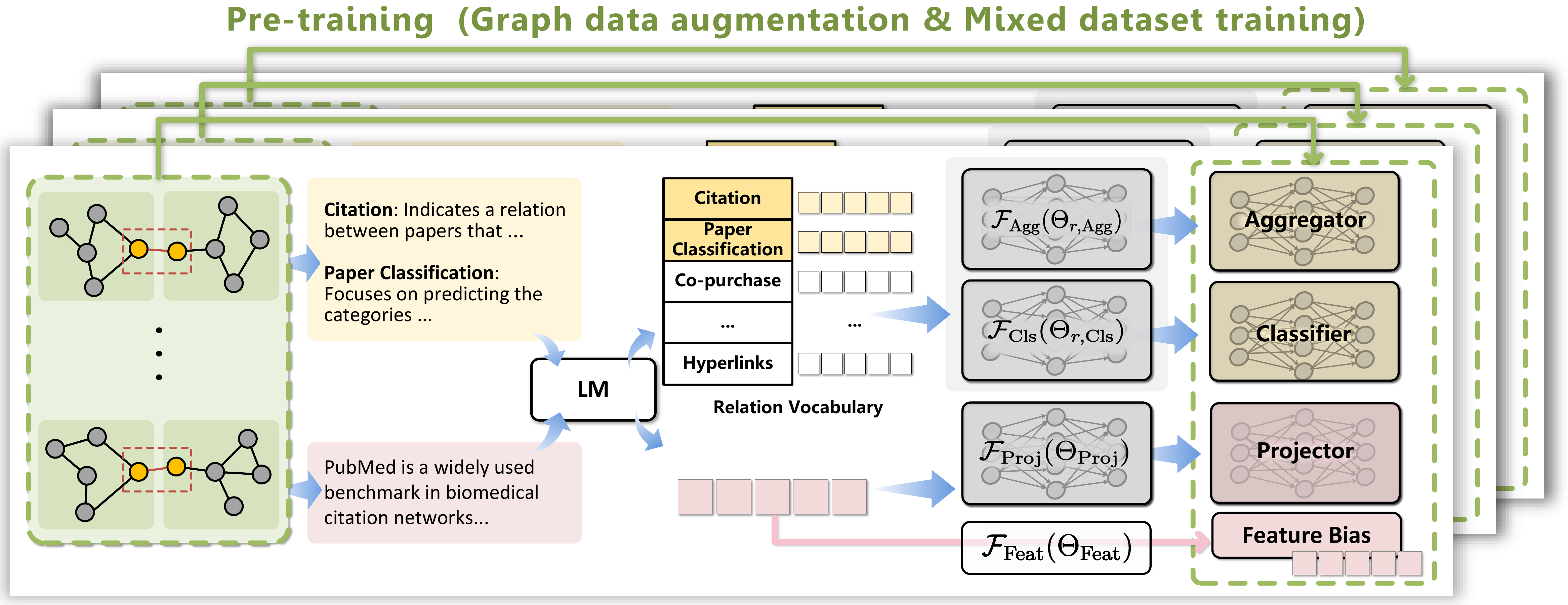}
    \caption{
    The overall framework of~\model.
}
    \label{fig:framework}
\end{figure*}

\paragraph{Graph data.}
Each graph dataset is represented as
$\mathcal{G} = \{\mathcal{V}_{\mathcal{G}},\mathcal{E}_{\mathcal{G}}, \allowbreak X_{\mathcal{G}},  \mathcal{R}_{\mathcal{G}}, t_\mathcal{G}\} $, 
where $\mathcal{V}_{\mathcal{G}}$ denotes the set of nodes,
$\mathcal{E}_{\mathcal{G}}$ represents the set of edges, 
$X_{\mathcal{G}}$ is the feature matrix corresponding to the nodes, 
$\mathcal{R}_{\mathcal{G}}$ is the collection of textual descriptions for different relation types,
and $t_\mathcal{G}$ indicates the text description of the dataset.
We define 
$ \mathcal{G} \in 
\mathcal{G}_{\text{source}} \cup 
\mathcal{G}_{\text{target}}$, 
where $\mathcal{G}_{\text{source}}$ represents the pretraining graph datasets,
$\mathcal{G}_{\text{target}}$ denotes the target graph datasets for transfer learning, 
and $\mathcal{G}_{\text{source}} \cap 
\mathcal{G}_{\text{target}} = \emptyset$.
Additionally,
we define relation vocabulary as $\mathcal{R}_{\text{source}}=\{r | r \in \mathcal{R}_{\mathcal{G}}, \mathcal{G} \in \mathcal{G}_{\text{source}}\}$,
where $r$ represents a relation token,
i.e.,
the textual description of the relation.

\paragraph{Objective.}
The primary objectives of this work are to evaluate two fundamental aspects of Graph Foundation Models (GFMs):
(1) \emph{Generalizability:} 
This aspect focuses on determining whether the model can effectively capture and leverage knowledge acquired from multiple pretraining datasets, enabling it to generalize well across diverse graph structures.
(2) \emph{Transferability:}
This evaluates the model’s ability to transfer the knowledge learned during pretraining to new, unseen datasets, and its effectiveness in adapting to different domains.

\paragraph{Pre-training Tasks (Relation Token Prediction).}
For each node $v_i$,
we construct a $k$-hop subgraph $s_i$
and define training triplets in the form
$\langle s_i, r_{ij}, s_j \rangle$.
Depending on the downstream task, these triplets fall into two categories:
(1) Classification tasks: 
$s_i$ denotes the subgraph centered at the node $v_i$ to be classified, while $s_j$ corresponds to a label node, represented as a subgraph containing only the label node itself. 
Each label node is defined as the centroid of all training samples associated with that specific label.
The relation token $r_{ij}$ denotes the textual description of the classification task within this domain.
(2) Link prediction tasks: $s_i$ and $s_j$ are the subgraphs of nodes $v_i$ and $v_j$, respectively, and $r_{ij}$ represents the textual description of the edge type between them.
For both tasks, $r_{ij} \in \mathcal{R}_{source}$ represents the potential relationship between the subgraphs $s_i$ and $s_j$. 
Our pretraining objective is to train a model $\mathcal{F}( {\Theta})$ to perform binary classification for different relations,
aiming to predict the following probability:
\begin{equation}
   P(y = 1 \mid \langle s_i, r_{ij}, s_j \rangle) = \mathcal{F}( \langle s_i, r_{ij}, s_j \rangle; \Theta) \ ,
\end{equation}
where $P(\cdot \mid \cdot)$ denotes the conditional probability that the relation $r_{ij}$ holds between subgraphs $s_i$ and $s_j$.
Thus, the goal of $\mathcal{F}$ is to predict whether this relation exists.

\section{Methodology}
In this section, we present \model, whose overall architecture is shown in Figure~\ref{fig:framework}. 
We first use LMs to generate relational and dataset representations based on their descriptions. 
These representations are mapped to the parameters of the aggregator and classifier through two hypernetworks, $\mathcal{F}_{\text{Agg}}$ and $\mathcal{F}_{\text{Cls}}$, respectively. 
The dataset representation is mapped to a feature projector via another hypernetwork $\mathcal{F}_{\text{Proj}}$, and a dataset-specific feature bias is derived 
to enhance flexibility. Finally, all components are pre-trained with graph data augmentation and a mixed-dataset strategy.

\subsection{Feature Alignment}

Since the node feature matrices across datasets may differ in dimension, we align them using the following transformation:
\begin{equation}
\label{eq:svd}
    \hat{x}_i = {\mathcal{T}}(x_i) \in \mathbb{R}^{d_{x}}, 
    \quad x_i \in \mathbb{R}^{d_{x_i}} \ ,
\end{equation}
where $x_i$ denotes the initial feature corresponding to node $v_i$,
$d_{x_i}$ and $d_x$ represent the initial attribute dimension and the transformed attribute dimension,
respectively.
The function 
$\mathcal{T}(\cdot)$ serves as a transformation operator that aligns feature representations across different datasets.
To ensure compatibility with graph data containing numerical attributes, 
we employ singular value decomposition (SVD) for this transformation.
For comparison, we also investigate LM-based feature initialization in Section~\ref{sec:lm_init}.

\subsection{Multi-Relation Hypernetwork Learning}

For each relation token $r$ in the relation vocabulary, we initialize its representation by using a language model (LM). 
Specifically, we use Sentence-BERT~\citep{sentencebert}, a widely used model for generating effective sentence representations. The initial relation representation for each token is denoted as:
\begin{equation}
h_r = \text{LM}(r), \quad r \in \mathcal{R}_{\text{source}}
\end{equation}

\subsubsection{Aggregator for Relation Modeling}
Inspired by the Relational Graph Convolutional Network (RGCN)~\citep{rgcn}, 
we propose a method that designs tailored aggregators for different relations,
accounting for the varying impacts of relationships on nodes.
Specifically, \model~adaptively adjusts the message passing mechanism with a hypernetwork that generates relation-specific aggregator parameters from relation representations; that is, the semantic information encoded in the language model is mapped into the aggregator parameters. 
The mapping can be formalized as follows:
\begin{equation}
\Phi_{r}^{(l)} = 
\mathcal{F}_{\text{Agg}}(h_r;\Theta_{r,\text{Agg}}^{(l)}), 
\quad l=1,2\dots,L \ ,
\end{equation}
where $\Phi_r^{(l)}$ denotes the aggregator parameters corresponding to relation token $r$ at the $l$-th layer of the graph neural network. 
The function $\mathcal{F}_{\text{Agg}}$ is a hypernetwork conditioned on the relation representation $h_r$, and parameterized by $\Theta_{r,\text{Agg}}^{(l)}$. 
 $L$ denotes the total number of layers in the graph neural network.
With the parameters of the aggregators obtained, 
we can perform message passing for different relations within the subgraph. 
In subgraph $s_i$, 
the representation $h_{v_i}$ of node $v_i$ at layer $l+1$ is computed as follows:
\begin{equation}
\label{eq:gnn}
h_{v_i}^{(l+1)} = 
h_{v_i}^{(l)} + \sum_{r \in \mathcal{R_{\text{source}}}} \sum_{v_j \in \mathcal{N}_i^r} \frac{1}{ | \mathcal{N}_i^r |} \Phi_r^{(l)} h_{v_j}^{(l)} \ ,
\end{equation}
where 
$\mathcal{N}_i^r$ represents the set of neighbors of node $v_i$ connected through relation token $r$.
To ensure balanced contributions from neighbors, 
we apply the normalization factor $\frac{1}{|\mathcal{N}_i^r|}$, mitigating the potential over-representation of densely connected nodes.
Finally, the representation of node $v_i$ from the last layer, $h_{v_i}^{(L)}$, serves as the overall representation $h_{s_i}$ for the corresponding subgraph $s_i$.

\subsubsection{Task-Specific Classifier for Relational Inference}
For each training triple $\langle s_i, r_{ij}, s_j \rangle$, we compute the representations of the two subgraphs, $h_{s_i}$ and $h_{s_j}$, respectively. 
Based on these subgraph representations,
\model~utilizes relation token $r_{ij}$ as a supervision signal to predict the type of relationship between the two subgraphs. 
Since different types of relations correspond to distinct semantics and mapping spaces, 
we design independent classifiers for each relation to determine whether a specific relation token $r$ exists between the subgraphs.
To effectively distinguish various types of relations,
we introduce another hypernetwork to adaptively learn the parameters of different classifiers based on the relation representation $h_r$.
The relation-specific classifier is defined as follows:
\begin{equation}
\Psi_{r} = \mathcal{F}_{\text{Cls}}(h_r;\Theta_{r,\text{Cls}}) \ ,
\end{equation}
where $\mathcal{F}_{\text{Cls}}(h_r;\Theta_{r,\text{Cls}})$ denotes the mapping function that takes the relation representation $h_r$ as input and outputs the classifier parameters $\Psi_{r}$.
Here, $\Theta_{r,\text{Cls}}$ represents the hypernetwork parameters specific to the classifier for relation token $r$.
Subsequently, 
the representations of subgraphs
$h_{s_i}$ and $h_{s_j}$ are combined via the Hadamard product ($\circ$) to form a joint representation $h_{s_{ij}}$.
The relation-specific classifier, parameterized by $\Psi_r$, is then applied to predict the probability $ P(y \mid \langle s_i, r_{ij}, s_j \rangle) $ of the given relation $r$ between subgraphs $s_i$ and $s_j$,
as described by the following equation:
\begin{equation}
\begin{aligned}
  h_{s_{ij}} &=  h_{s_i} \circ h_{s_j}, \\
  P(y \mid \langle s_i, r_{ij}, s_j \rangle) &=  \text{sigmoid}(\Psi_r h_{s_{ij}}) \in (0,1) \ .
\end{aligned}
\end{equation}

Based on the two aforementioned hypernetworks, 
\model~constructs mapping functions that transform relation tokens into parameters of the aggregators and classifiers. 
This design allows the graph foundation model to effectively capture a wide range of relations during pretraining, 
thereby enabling the development of specialized aggregators and classifiers tailored to each specific relation. 
REEF thus possesses the ability to generalize across diverse relation scenarios.

\subsection{Feature Representation Enhancement}
In the process of designing the aggregators and classifiers,
we observe that different datasets 
exhibiting
the same relation can share the same aggregator and classifier parameters.
However, node attributes across different datasets exhibit significant variations, reflecting the distinct characteristics inherent to each dataset.
For example, both Cora and PubMed are citation networks where the edges represent \emph{citation relationships}, and their classification tasks focus on \emph{research papers}.
Despite this similarity,
the node attributes differ significantly:
Cora predominantly contains papers related to \emph{computer science}, while PubMed is centered around \emph{diabetes research}.
Such differences make it insufficient to only condition on relations;
the distinct characteristics of each dataset must also be captured.
To this end, 
REEF 
learns a dedicated dataset representation and adapts it through two complementary mechanisms: a dataset-specific projector and a dataset-level feature bias.

Specifically,
based on the dataset description $t_\mathcal{G}$ of dataset $\mathcal{G}$, 
we encode it using Sentence-BERT to obtain its representation $h_{t_\mathcal{G}}$.
Here,
$h_{t_\mathcal{G}}$ is a learnable embedding that is updated during training,
thereby
enhancing the flexibility and expressiveness of the representation.
Then,
we apply the mapping function $\mathcal{F}_{\text{Proj}}$, implemented by the hypernetwork, to generate the projection matrix,
which is formulated as follows:
\begin{equation}
h_{t_{\mathcal{G}}}= {\text{LM}}(t_{\mathcal{G}}), \quad
\Phi_{\mathcal{G}}^{(l)} = \mathcal{F}_{\text{Proj}}(h_{t_{\mathcal{G}}};\Theta_{\mathcal{G},\text{Proj}}^{(l)}) \ , 
\end{equation}
where 
$\Phi_{\mathcal{G}}^{(l)}$ and $\Theta_{\mathcal{G},\text{Proj}}^{(l)}$ 
represent the parameters of the $l$-th layer, with the former being for the projector and the latter for the hypernetwork.
The projection matrix 
is then applied directly to the representation of the target node $v_i$ at each layer, adapting its hidden state to dataset-specific characteristics.
Accordingly,
the update of node $v_i$ at layer $l+1$ in the GNN,
as given in 
Eq.~\ref{eq:gnn},
is reformulated as:
\begin{equation}
h_{v_i}^{(l+1)} = 
\textcolor[HTML]{205867}{\Phi_{\mathcal{G}}^{(l)}}
h_{v_i}^{(l)} 
+ \sum_{r \in \mathcal{R_{\text{source}}}} \sum_{v_j \in \mathcal{N}_i^r} \frac{1}{ | \mathcal{N}_i^r |} \Phi_r^{(l)} h_{v_j}^{(l)} \ ,
\end{equation}

Furthermore, 
we incorporate a dataset-level feature bias to the initial representation of all nodes in the subgraph.
This bias is derived from $h_{t_\mathcal{G}}$ through a transformation function $\mathcal{F}_{\text{Feat}}$ parameterized by $\Theta_{\text{Feat}}$:
\begin{equation}
 h_{\mathcal{G}}=
\mathcal{F}_{\text{Feat}}(h_{t_{\mathcal{G}}};\Theta_{\text{Feat}}) \in \mathbb{R}^{d_{x}},
\quad
h_{v}^{(0)} = \hat{x}+h_{\mathcal{G}}, \; \forall v \in s_i\ 
\end{equation}
where $\hat{x}$ represents the transformed feature of node $v$ as defined in Eq.~\ref{eq:svd}.
This bias term $h_{\mathcal{G}}$ provides a global dataset-level shift, aligning node features with the broader semantic distribution of the dataset (e.g., shifting toward biomedical semantics in PubMed).
Together, these two mechanisms allow REEF to balance local flexibility with global consistency, leading to more robust representation learning across diverse datasets.

In summary, 
the parameterized hypernetworks in REEF include
$\{ \Theta_{r,\text{Agg}}, \Theta_{r,\text{Cls}}, \Theta_{\mathcal{G},\text{Proj}} \}$,
where $r\in\mathcal{R_{\text{source}}}$.
For each relation token $r$,
the parameters of its tailored aggregator and classifier, denoted as
$\{\Phi_r^{(1)},\Phi_r^{(2)},\dots, \allowbreak
\Phi_r^{(L)}\}$ and $\Psi_r$
are computed by two hypernetworks
$\mathcal{F}_{\text{Agg}}$ and $\mathcal{F}_{\text{Cls}}$,
respectively.
The hypernetwork
$\mathcal{F}_{\text{Proj}}$
generates
the parameters of dataset projector $\{\Phi_{\mathcal{G}}^{(1)},\Phi_{\mathcal{G}}^{(2)},\dots, \Phi_{\mathcal{G}}^{(L)} \}$.
Moreover, the feature bias transformation is achieved by learning parameters $\Theta_{\text{Feat}}$ in $\mathcal{F}_{\text{Feat}}$.
Building upon this framework, we proceed with the pre-training of the graph foundation model.

\subsection{Pre-training Strategy}
To enhance the robustness and generalization capabilities of~\model,
we introduce
graph data augmentation 
and a mixed dataset training strategy during pretraining.
These techniques are designed to capture relational diversity more effectively, 
mitigate overfitting, 
and enhance adaptability across various graph structures.
Specifically, graph data augmentation introduces edge perturbation by randomly removing edges in each training epoch, encouraging the model to accommodate topological variations. 
In parallel, 
\model~is trained using a mixed-dataset strategy across multiple graph domains.
Each dataset is divided into batches, and all batches are randomly shuffled to construct a unified training queue. 
This design not only mitigates the risk of overfitting to a single dataset but also balances datasets of varying scales,
thereby enhancing generalization.

For downstream tasks,
the pretrained foundation model is fine-tuned 
on the target dataset $\mathcal{G} \in \mathcal{G_{\text{target}}}$.
The corresponding aggregators and classifiers are obtained by
the learned relation tokens from the relation vocabulary.
In addition, 
the dataset representation $h_{t_\mathcal{G}}$
is learned specifically for $\mathcal{G}$.

\section{Experiments}
\label{sec:exp}

We conduct comprehensive experiments to evaluate REEF from multiple perspectives:

\begin{itemize}[leftmargin=1.5em, itemsep=1pt, parsep=0pt, topsep=0.2pt, partopsep=0pt]
    \item \textbf{RQ1:} Can REEF learn generalizable knowledge through multi-domain pretraining?
    
    \item \textbf{RQ2:} Can REEF effectively transfer relational knowledge under few-shot settings?
    
    \item \textbf{RQ3:} Can REEF generalize across diverse graph domains with substantially different structures and semantics?
    
    \item \textbf{RQ4:} Can REEF generalize to completely unseen domains and tasks in a zero-shot manner?

     \item \textbf{RQ5:} How does REEF transfer across different feature modalities with LM-based node features?

      \item \textbf{RQ6:} How do the main components of~\model~contribute to its overall performance?
\end{itemize}

Due to space limitation, we defer additional results to Appendix, including
performance on link prediction tasks (Appendix~\ref{app:lp}, \textbf{RQ7}), scaling laws across different dataset scales (Appendix~\ref{app:scaling}, \textbf{RQ8}), transfer performance on out-of-domain relations (Appendix~\ref{app:transfer_outdomain}, \textbf{RQ9}), the impact of varying hidden dimensions (Appendix~\ref{app:hyper_hidden}, \textbf{RQ10}), visualization of learned representations (Appendix~\ref{app:visual}, \textbf{RQ11}), computational complexity analysis (Appendix~\ref{app:complex}, \textbf{RQ12}), performance trends during pre-training (Appendix~\ref{app:trends}, \textbf{RQ13}), comparison of aggregator designs (Appendix~\ref{app:agg_compare}, \textbf{RQ14}), the effect of nonlinear activations (Appendix~\ref{app:activation}, \textbf{RQ15}), empirical efficiency analysis (Appendix~\ref{app:efficiency}, \textbf{RQ16}), comparison with heterophily-specific GNNs (Appendix~\ref{app:heterophily}, \textbf{RQ17}), and extension to heterogeneous graphs (Appendix~\ref{app:hetero_graph}, \textbf{RQ18}).
Together, these evaluations comprehensively examine the generalizability, transferability, robustness, and scalability of REEF under substantially different graph domains and task settings.

\begin{table*}[t]
    \centering
    \caption{Comparison of model performance on various pretraining datasets. 
    The table reports accuracy (Acc $\%$) and the performance gap $(\Delta = \text{Top} - \text{Score})$ to the best result on each dataset. 
    The average ``Acc'' is the mean accuracy calculated across the five datasets presented in the table.
    The \underline{\textbf{best}} result is bold and underlined, while the \underline{runner-up} is underlined.
    }
    
    \resizebox{0.98\textwidth}{!}{
    \begin{tabular}{l|ccccc|cc}
        \toprule
        
        \textbf{Domain} & \multicolumn{2}{c}{\textbf{Citation}} & \multicolumn{2}{c}{\textbf{WebKB}} & \textbf{Amazon} & \multicolumn{2}{c}{\textbf{Average}} \\
        \cmidrule(lr){1-1}
        \cmidrule(lr){2-3} \cmidrule(lr){4-5} \cmidrule(lr){6-6} \cmidrule(lr){7-8}
        \textbf{Dataset} & \textbf{Pubmed} & \textbf{Citeseer} & \textbf{Wisconsin} & \textbf{Texas} & \textbf{Photo} & \textbf{Acc $\uparrow$} & \textbf{$\Delta \downarrow$} \\
        \midrule
        GCN (ind) & \textbf{\underline{88.42}} ($\Delta_{0.00}$) & \underline{76.50} ($\Delta_{0.05}$) & 51.76 ($\Delta_{24.71}$) & 55.14 ($\Delta_{25.81}$) & \textbf{\underline{93.02}} ($\Delta_{0.00}$) & 72.97 & 10.11 \\
        GAT (ind) & 86.33 ($\Delta_{2.09}$) & \textbf{\underline{76.55}} ($\Delta_{0.00}$) & 49.41 ($\Delta_{27.06}$) & 52.16 ($\Delta_{28.79}$) & 92.73 ($\Delta_{0.29}$) & 71.44 & 11.65 \\
        GCN (joint) & 61.21 ($\Delta_{27.21}$) & 64.86 ($\Delta_{11.69}$) & 61.21 ($\Delta_{15.26}$) & 57.14 ($\Delta_{23.81}$) & 30.45 ($\Delta_{62.57}$) & 54.97 & 28.11 \\
        GAT (joint) & 52.93 ($\Delta_{35.49}$) & 57.81 ($\Delta_{18.74}$) & 55.81 ($\Delta_{20.66}$) & 52.38 ($\Delta_{28.57}$) & 28.52 ($\Delta_{64.50}$) & 49.49 & 33.59 \\
        RGCN (joint) & 80.30 ($\Delta_{8.12}$) & 71.77 ($\Delta_{4.78}$) & \underline{72.09} ($\Delta_{4.38}$) & \textbf{\underline{80.95}} ($\Delta_{0.00}$) & 79.03 ($\Delta_{13.99}$) & \underline{76.83} & \underline{6.25} \\
        \midrule
        REEF & \underline{86.38} ($\Delta_{2.04}$) & 72.37 ($\Delta_{4.18}$) & \textbf{\underline{76.47}} ($\Delta_{0.00}$) & \underline{70.27} ($\Delta_{10.68}$) & \underline{93.01}($\Delta_{0.01}$) & \textbf{\underline{79.70}} & \textbf{\underline{3.38}} \\
        \bottomrule
    \end{tabular}
    }
    \label{tab:pre}
\end{table*}

\begin{table*}[t]
    \centering
    \caption{Performance comparison of different methods under transfer learning (1-shot).
    ``GCOPE+CL'' and ``GCOPE+Sim'' denote the variants of GCOPE that utilize the contrastive learning frameworks of GraphCL and SimGRACE, respectively.
    The \textbf{best} results are highlighted in bold.
    ``Improve.'' represents the improvement of \model~over the best baseline on the corresponding metric.
    }
    \label{tab:transfer}
    \resizebox{1.0\textwidth}{!}{
    \begin{tabular}{l|ccc|ccc|ccc}
        \toprule
        \multirow{2}{*}{Method} & \multicolumn{3}{c}{Cora (Citation)} & \multicolumn{3}{c}{Cornell (WebKB)} & \multicolumn{3}{c}{Computers (Amazon)} \\
        \cmidrule(lr){2-4} \cmidrule(lr){5-7} \cmidrule(lr){8-10}
         & Acc & AUC & F1 & Acc & AUC & F1 & Acc & AUC & F1 \\
        \midrule
        GCN & $0.3012_{\pm.06}$ & $0.6444_{\pm.04}$ & $0.2591_{\pm.04}$ & $0.3263_{\pm.04}$ & $0.5666_{\pm.01}$ & $0.3151_{\pm.03}$ & $0.2602_{\pm.07}$ & $0.6773_{\pm.02}$ & $0.2428_{\pm.04}$ \\
        GAT & $0.3646_{\pm.04}$ & $0.6769_{\pm.03}$ & $0.3108_{\pm.04}$ & $0.3275_{\pm.14}$ & $0.5306_{\pm.03}$ & $0.1497_{\pm.04}$ & $0.3482_{\pm.07}$ & $0.6878_{\pm.05}$ & $0.2397_{\pm.05}$ \\
        GraphCL & $0.2507_{\pm.06}$ & $0.6350_{\pm.03}$ & $0.2240_{\pm.03}$ & $0.4175_{\pm.04}$ & $0.6350_{\pm.02}$ & $0.3500_{\pm.04}$ & $0.2856_{\pm.04}$ & $0.6467_{\pm.03}$ & $0.1653_{\pm.06}$ \\
        SimGRACE & $0.2492_{\pm.02}$ & $0.5765_{\pm.03}$ & $0.1567_{\pm.04}$ & $0.3438_{\pm.13}$ & $0.5954_{\pm.09}$ & $0.2168_{\pm.09}$ & $0.2666_{\pm.10}$ & $0.6286_{\pm.01}$ & $0.1603_{\pm.03}$ \\
        GCOPE+CL & $0.3368_{\pm.02}$ & $0.6971_{\pm.04}$ & $0.2967_{\pm.03}$ & $0.3975_{\pm.10}$ & $0.6694_{\pm.04}$ & $0.3120_{\pm.04}$ & $0.3439_{\pm.03}$ & $0.7023_{\pm.01}$ & $0.2976_{\pm.03}$ \\
        GCOPE+Sim & $0.2525_{\pm.05}$ & $0.5744_{\pm.03}$ & $0.1722_{\pm.06}$ & $0.3675_{\pm.09}$ & $0.6045_{\pm.04}$ & $0.2339_{\pm.04}$ & $0.3230_{\pm.01}$ & $0.6994_{\pm.00}$ & $0.2515_{\pm.00}$ \\
        MDGPT & $0.4421_{\pm.08}$ & $0.7912_{\pm.05}$ & $0.4271_{\pm.08}$ & $0.2400_{\pm.08}$ & $0.5776_{\pm.04}$ & $0.1723_{\pm.04}$ & $0.3837_{\pm.09}$ & $0.7950_{\pm.04}$ & $0.4120_{\pm.08}$ \\
        \midrule
        \textbf{\model} & $\textbf{0.4866}_{\pm.04}$ & $\textbf{0.8035}_{\pm.03}$ & $\textbf{0.4800}_{\pm.05}$ & $\textbf{0.4865}_{\pm.08}$ & $\textbf{0.7016}_{\pm.03}$ & $\textbf{0.3649}_{\pm.07}$ & $\textbf{0.4772}_{\pm.06}$ & $\textbf{0.8480}_{\pm.03}$ & $\textbf{0.4403}_{\pm.02}$ \\
        Improve. & $+10.07\%$ & $+1.55\%$ & $+12.39\%$ & $+16.52\%$ & $+4.81\%$ & $+4.08\%$ & $+24.05\%$ & $+6.67\%$ & $+6.87\%$ \\
        \bottomrule
    \end{tabular}
    }
\end{table*}

\subsection{Main Experimental Setup}

\textbf{Datasets.}
For a fair comparison, we conduct experiments on real-world datasets from four different domains,
including (1) two knowledge graphs: FB15K237 and WN18RR~\citep{toutanova2015observed}; (2) three citation networks: Pubmed, Citeseer, and Cora~\citep{sen2008collective,yang2016revisiting}; (3) three WebKB subsets: Wisconsin, Texas, and Cornell~\citep{pei2020geom}; and (4) two Amazon co-purchase networks: Computers and Photo~\citep{shchur2018pitfalls}.
Detailed descriptions on these datasets are provided in 
Appendix~\ref{app:datasets}.

\textbf{Baselines.}
We compare our proposed framework with three groups of baseline models:
(1) 
Supervised GNN methods
including GCN~\citep{gcn}, GAT~\citep{gat}, and RGCN~\citep{rgcn}.
(2) Graph contrastive learning methods like GraphCL~\citep{you2020graph} and SimGRACE~\citep{xia2022simgrace}.
(3) Graph pre-training methods for transfer: GCOPE~\citep{zhao2024all} and
MDGPT~\citep{yu2024text}.
For further details, refer to Appendix~\ref{app:baselines}.

These datasets and baselines are used in the pretraining and few-shot transfer learning tasks.
In the experiment,
we perform comprehensive study to measure the generalizability of \model.
However,
most state-of-the-arts are either only applicable to part of scenarios
or have not reported their results in some tasks.
Therefore, 
we select different representative state-of-the-arts for various tasks and perform comparison in their experimental settings. In this way, we can directly report their published results to ensure fair comparison.

\subsection{Pre-training Performance (RQ1)}

REEF is pretrained on seven graph datasets spanning multiple domains, including knowledge graphs, citation networks, and heterophilous web graphs. Relation tokens are constructed from edge semantics or task descriptions associated with each dataset. 
Implementation details are given in Appendix~\ref{app:exp_setting}.

Table~\ref{tab:pre}
presents the pre-training experimental results of~\model\ compared to other baseline models. 
For baselines, 
``ind'' indicates models trained and tested independently on a single dataset, 
while ``joint'' represents models trained simultaneously on five datasets: 
Pubmed, Citeseer, Wisconsin, Texas, and Photos.
In the ``joint'' setting, 
all methods utilize the same input features as~\model, 
and the labels from all datasets are combined into a unified classification space. 
i.e., 
the number of output categories is equal to the sum of the categories of all datasets.
Furthermore, in RGCN, datasets from the same domain share identical edge types.

From the table, we draw the following key findings:
(1) 
\model~achieves the best overall performance. 
Among all methods, 
\model~attains the highest average accuracy of 79.70\%, demonstrating superior overall performance and maintaining a leading edge across multiple datasets.
(2)
\model~demonstrates more stable performance across different datasets,
consistently achieving near-optimal or best accuracy.
For example,
it attains 76.47\% on Wisconsin, 70.27\% on Texas, and 93.01\% on Photo.
In contrast, other methods show significant performance fluctuations across different datasets. 
Notably, 
\model~has the lowest performance gap ($\Delta$=3.38) among all methods, 
indicating minimal variance across datasets and the highest overall stability.
(3)
\model~and RGCN significantly outperform GCN and GAT in the joint training setting,
showing that incorporating relation semantics enhances cross-domain generalization.
(4)
Compared with RGCN, REEF further benefits from its relation vocabulary and dataset-specific hypernetworks, which tailor aggregators and classifiers to relation semantics ($\mathcal{F}_{\text{Agg}}, \mathcal{F}_{\text{Cls}}$) while distinguishing dataset-specific variations via projectors ($\mathcal{F}_{\text{Proj}}$). This design enables both relation-level and dataset-level adaptation, resulting in stronger generalization.

In conclusion, \model~demonstrates outstanding generalization, significantly outperforming other methods due to the universal knowledge learned from relation tokens and enhanced dataset-level feature representations.

\subsection{Transfer Learning Performance (RQ2)}
We further compare the transfer performance of the pretrained model
across datasets from different domains. 
For 
graph contrastive learning methods,
the backbone is GCN, 
and the results of the baseline models are directly reported from~\citep{zhao2024all}.
The overall results are presented in Table~\ref{tab:transfer}.
From the table, we observe the following:
(1) 
\model~demonstrates significantly superior transfer performance across all datasets, achieving substantial improvements on all three metrics,
with particularly remarkable gains on the Computers dataset, where REEF improves accuracy by 24.05\%, AUC by 6.67\%, and F1 score by 6.87\%.
(2) Compared to GCOPE and MDGPT that treat the dataset level as the fundamental unit,
REEF places greater emphasis on predicting relationships between nodes, thereby enabling more coherent and effective knowledge transfer across domains.
(3) These results underscore the critical use of relation tokens in improving the generalizability of graph foundation models.

\begin{table}[t]
\centering
\caption{
Performance comparison across different graph domains.
Best results are highlighted in bold.
Detailed results for each individual dataset are provided in Table~\ref{tab:comparison_diverse} of Appendix~\ref{app:complete_results}.
}
\resizebox{0.9\textwidth}{!}{
\begin{tabular}{lccccc}
\toprule
Domain & \#Datasets & GAT & GraphAny & TS-Mean & REEF \\
\midrule

Citation Networks & 4
& 64.23 & 70.14 & 63.39 & \textbf{71.08} \\

Co-authorship Networks & 4
& 77.63 & 81.82 & 81.03 & \textbf{83.00} \\

Social Networks & 6
& 69.00 & 69.03 & 68.31 & \textbf{70.25} \\

Co-purchase Graphs & 3
& 64.12 & \textbf{71.58} & 71.27 & 71.00 \\

Heterophilic Benchmarks & 6
& 72.85 & 69.34 & 68.53 & \textbf{76.77} \\

Air-traffic Graphs & 3
& 39.14 & 40.38 & 39.15 & \textbf{42.64} \\

\midrule
Average (\%) $\uparrow$
& 26
& 66.47 & 68.23 & 66.54 & \textbf{70.75} \\

\bottomrule
\end{tabular}
}
\label{tab:comparison_domain_diverse_simple}
\end{table}

\subsection{Cross-Domain Generalization Across Diverse Graph Domains (RQ3)}
\label{sec:graphany}

To further evaluate the cross-domain generalizability of REEF, we conduct a large-scale transfer experiment across 26 graph datasets spanning substantially different domains.
We follow the evaluation protocol of GraphAny~\cite{zhao2024graphany}, where all methods use the same dataset splits for fair comparison.
We compare REEF with representative graph learning baselines, including GAT~\cite{gat}, GraphAny~\cite{zhao2024graphany}, and TS-Mean~\cite{finkelshtein2025equivariance}. 
Detailed dataset categorization and experimental settings are provided in Appendix~\ref{app:complete_results}.

Table~\ref{tab:comparison_domain_diverse_simple} reports the averaged performance across six representative graph domains.
REEF achieves the best overall performance among all methods, demonstrating strong generalizability across substantially different graph structures and semantic distributions.
Notably, REEF shows the largest improvement on Heterophilic Benchmarks, suggesting that relation-token modeling helps alleviate the challenges caused by diverse neighborhood semantics.
REEF also performs particularly well on Air-traffic Graphs, where relational interactions between nodes play a central role.
These results indicate that REEF learns transferable relational knowledge that generalizes effectively across diverse graph domains with different topological properties and feature characteristics.

\subsection{Zero-Shot Transfer to Unseen Domain and Task (RQ4)}

To evaluate the cross-domain and cross-task generalizability of REEF, we follow the experimental setting of GOFA~\cite{kong2024gofa} and conduct zero-shot molecular graph classification on the BBBP and HIV benchmarks. 
For consistency and fairness, we directly report baseline results from the original GOFA paper and additionally include REEF under the same evaluation setting. 
``FT Scale'' denotes the number of question--answer pairs used for fine-tuning.
Note that REEF is evaluated on molecular graph classification, a task entirely unseen during training, with molecular datasets absent from the pretraining corpus, providing a strict cross-task and cross-domain setting. AUROC is used as the evaluation metric.

\begin{wraptable}{r}{0.50\textwidth}
\vspace{-0.5em}
\centering
\caption{
Zero-shot transfer performance on molecular graph benchmarks without molecular-domain pretraining or fine-tuning.
}
\label{tab:zero_shot}


\begin{tabular}{lccc}
\toprule
Method & FT scale & BBBP & HIV \\
\midrule
OFA~\cite{liu2023one} & N/A & -- & 35.67 \\
MoMu~\cite{su2022molecular} & N/A & 49.81 & 50.26 \\
Galactica~\cite{taylor2022galactica} & N/A & 53.94 & 33.85 \\
GIMLET~\cite{zhao2023gimlet} & 400M QA & 59.39 & 66.24 \\
GOFA~\cite{kong2024gofa} & 100k QA & 54.91 & 53.02 \\
\midrule
\model & \textbf{None} & 54.80 & 63.17 \\
\bottomrule
\end{tabular}

\end{wraptable}

Table~\ref{tab:zero_shot} reports the zero-shot transfer performance.
REEF achieves competitive performance on BBBP and outperforms GOFA by approximately 10\% on HIV.
Notably, this strong transfer performance is attained with substantially weaker supervision and without access to molecular graph data during pretraining. However, GOFA is fine-tuned on over 100k ChemBL question–answer pairs, while GIMLET is pretrained on more than 360k molecular graphs.
These findings show that REEF learns highly transferable relational representations that enable robust zero-shot generalization across unseen domains  and tasks, highlighting its potential as a general-purpose graph foundation model.

\begin{wraptable}{r}{0.5\textwidth}
\vspace{-0.6em}
\centering

\caption{
3-shot transfer learning results using LM-based node features.
$^\dagger$ denotes that no E-commerce graph data is used during pretraining, making History and Ratings cross-domain transfer datasets.
The best and second-best results are highlighted in \textbf{bold} and \underline{underline}, respectively.
}

\label{tab:transfer_lm_based}

\setlength{\tabcolsep}{4pt}

\begin{tabular}{lccc}
\toprule
Method & Cora & History & Ratings \\
\midrule
GraphMAE & 72.49 & 39.15 & 31.68 \\
LLaGA & 60.70 & 36.45 & 23.45 \\
OFA & 52.49 & 39.36 & 29.08 \\
OFA-FS & 42.10 & 17.50 & 20.50 \\
Prodigy (M) & 23.40 & 12.71 & 20.16 \\
Prodigy & 40.59 & 19.47 & 20.84 \\
GraphText & 50.33 & 48.00 & 37.67 \\
GraphAdapter & 33.74 & 36.60 & 29.41 \\
GraphLLM & 68.00 & 56.67 & 34.33 \\
GPPT & 44.14 & 27.54 & 14.24 \\
Gprompt & 55.38 & 17.36 & 15.38 \\
SBERT (w/o PT) & 68.42 & 51.25 & 20.95 \\
 GIT-G &  60.93 &50.78 &  21.21    \\
GIT-S & 63.23 & 50.99  & 21.17  \\
\midrule
REEF (S1)$^\dagger$ & \textbf{75.95} & \underline{57.40} & \underline{37.20} \\
REEF (S2) & \underline{74.58} & \textbf{62.82} & \textbf{38.24} \\
\bottomrule
\end{tabular}

\end{wraptable}

\subsection{Transfer Learning with LM-based Node Features (RQ5)}
\label{sec:lm_init}

In the main experiments, node features are represented as numerical attributes.
To further evaluate the cross-feature generalizability of~\model, we additionally conduct transfer learning experiments using LM-based node features derived from textual attributes.
Following the evaluation protocol of TSGFM~\citep{tsgfm}, we evaluate REEF under a 3-shot transfer learning setting with only three labeled samples per class. 
Node textual attributes are encoded into embeddings using Sentence-BERT~\citep{sentencebert}. 
The evaluation covers citation-network and E-commerce graph datasets. 
REEF is pretrained under two configurations:
(i) \textit{Setting~1} (S1), using FB15K237, WN18RR, Pubmed, and Citeseer, without any E-commerce graph data during pretraining;
and
(ii) \textit{Setting~2} (S2), which further includes the Photo dataset from the E-commerce domain.
We compare REEF with representative 
baselines, including graph self-supervised methods, 
graph prompt learning method, LLM-based graph learning methods, and graph foundation models.
Detailed dataset descriptions, baseline configurations, and implementation details are provided in Appendix~\ref{app:lm_feature_transfer}.

As shown in Table~\ref{tab:transfer_lm_based}, REEF demonstrates consistently strong transferability across text-attributed graphs, highlighting its robustness across feature modalities and graph domains. REEF further outperforms several representative TAG-oriented baselines. 
For example, REEF achieves 75.95\% on Cora and 62.82\% on History, substantially outperforming the strongest competing baselines
(72.49\% and 56.67\%, respectively).
Notably, under Setting~1, the transfer datasets History and Ratings belong to the E-commerce domain, which is entirely absent from the pretraining corpus. Although Setting~1 performs slightly lower than Setting~2 on these datasets, it still surpasses most competing methods, further demonstrating the strong cross-domain transferability of REEF.

\begin{table*}[hbtp]
\centering
\caption{Ablation Study on~\model. ``Pre.'' refers to pretraining, ``Trans.'' refers to transfer learning, and ``All'' denotes the overall average performance.
We highlight the best score in bold.
}
\label{tab:ablation}
\resizebox{1.\textwidth}{!}{
\begin{tabular}{l|ccccccc|ccc|ccc}
\toprule
\multirow{2}{*}{\textbf{Method}} & \multicolumn{7}{c}{\textbf{Pre-training datasets}} & \multicolumn{3}{c}{\textbf{Transfer datasets}} & \multicolumn{3}{c}{\textbf{Avg.}}  \\
\cmidrule(lr){2-8} \cmidrule(lr){9-11} 
\cmidrule(lr){12-14}
& \textbf{FB15K237} & \textbf{WN18RR} &  \textbf{Pubmed} & \textbf{Citeseer} &  \textbf{Wisconsin} & \textbf{Texas}  & \textbf{Photo} & \textbf{Cora} & \textbf{Cornell} & \textbf{Computers} & \textbf{Pre.} & \textbf{Trans.} & \textbf{All} \\

\midrule        

REEF-LM  & \large \textbf{91.33}  & \large 94.06  & \large 86.05  & \large 70.42  &  \large 64.71  & \large 59.46  &  \large \textbf{93.53}  & \large 46.58  & \large 44.72  & \large 43.46  & \large 79.94  & \large 44.92  & \large 69.43 \\
REEF-FB & \large 89.90 & \large 92.18 & \large 86.23 & \large 70.27 & \large 70.59 & \large 54.05 & \large 92.16 & \large 40.07 & \large 44.51 & \large 42.09 & \large 79.34 & \large 42.22 & \large 68.21 \\
REEF-FP & \large 90.72 & \large 94.29 & \large 85.04 & \large 73.12 & \large 56.86 & \large 59.46 &  \large 93.33 & \large 40.22 & \large 43.69 & \large 46.50 & \large 78.97 & \large 45.10 & \large 68.32 \\
REEF-AGU & \large 91.14  & \large 93.78  & \large \textbf{86.41}  & \large 71.62 & \large 68.63    & \large 64.86  &  \large 92.22  & \large 40.81  & \large 43.69  & \large 46.68  & \large 81.24  & \large 43.73  & \large 69.98 \\
REEF     & \large 91.04  & \large \textbf{94.70}  & \large 86.38 & \large \textbf{72.37} & \large \textbf{76.47}  & \large \textbf{70.27}   & \large 93.01  & \large \textbf{48.66}  & \large \textbf{48.65}  & \large \textbf{47.72}  & \large \textbf{83.46}  & \large \textbf{48.34}  & \large \textbf{72.93} \\

\bottomrule
\end{tabular}
}

\end{table*}

\subsection{Ablation Study (RQ6)}
\label{app:ab_study}
We conduct an ablation study on~\model~to better
understand the contribution of its main components.
To assess the effectiveness of the relational representations in the relation vocabulary, 
we initialize the relational representations randomly, 
rather than obtaining them from LM,
referring to this variant as REEF-LM.
To investigate whether feature augmentation across different datasets under the same relations can further enhance the model’s capabilities, we train two variants without the feature bias and feature projector, named REEF-FB and REEF-FP, respectively.
Moreover, to evaluate the impact of graph data augmentation on the model’s generalization, 
we use a variant without graph data augmentation, called REEF-AGU.
By the way, 
we have validated 
the combined effect of the tailored aggregators, classifiers, and feature representation enhancement
in our experiments.
By comparing REEF with RGCN, 
the results demonstrate that the hypernetwork design enables the model to learn both the specific relational characteristics of each relation token and the dataset-specific distribution.

Table~\ref{tab:ablation} presents the results.
We can observe that:
(1) \model~
achieves the highest average score on pre-training datasets. 
Among the variants, 
REEF-LM performs the worst, 
indicating the crucial role of semantic information in relational representation initialization.
REEF-FB and REEF-FP perform worse overall,
indicating that the model lacks the ability to distinguish finer-grained variations across datasets.
REEF-AGU also shows a moderate decline, indicating that graph augmentation enhances the model’s generalizability.
(2) On transfer datasets, 
REEF again achieves the best performance, demonstrating its superior transferability. 
REEF-LM again performs relatively poorly, 
further confirming that the lack of LM-based relational representations, which provide crucial semantic information, leads to weaker knowledge transfer.
Removing feature prompts leads to a slight drop, reinforcing their role in improving transfer learning. 
The absence of graph augmentation results in a  decline, suggesting that while augmentation aids generalization. 
(3) Overall, 
\model~consistently surpasses its ablated counterparts, demonstrating that LM-based relation embeddings, feature prompts, and graph augmentation are all indispensable for achieving strong generalization and transferability.

\section{Conclusion and Future Work}
\label{sec:couclusion}
In this paper,
we proposed \model, a novel framework that introduces relation tokens as fundamental units for graph foundation models. 
We constructed a relation vocabulary to store relational information and design two hypernetworks to adaptively generate the parameters of aggregators and classifiers based on relation tokens.
Additionally,
another hypernetwork generates dataset-specific feature projectors and incorporate a dataset-level feature bias based on dataset description, 
improving flexibility across datasets with the same relations. 
Comprehensive experiments and analyses demonstrate that \model~performs exceptionally well in both pretraining and transfer learning tasks, indicating strong generalizability and transferability, and further revealing scaling-law behaviors to some extent.
These findings highlight the pivotal role of relations as fundamental units, suggesting that relation-centric design is essential for advancing the next generation of graph foundation models.

Looking ahead, we envision the relation-centric paradigm as a stepping stone toward truly universal graph models. Crucially, \model~treats relations as transferable semantic units \emph{without relying on large language models at inference}, offering a lightweight yet highly extensible foundation: enlarging the relation vocabulary to cover far more relation types and domains could enable a single pretrained model to transfer across an ever-broader spectrum of graphs. 
Building on this, promising directions include modeling higher-order and compositional relations, and extending to dynamic and temporal graphs, moving toward graph foundation models that generalize across tasks and domains as seamlessly as today's foundation models do for language and vision.





{
\small

\bibliographystyle{abbrv}
\bibliography{example_paper}



}


\appendix

\section{Related Work}

With the remarkable success of LLMs, GFMs have emerged as a prominent research direction in the graph domain~\citep{maoposition,dong2024universal,liu2023towards},
aiming to improve generalization across tasks and domains through training on large-scale and diverse graph datasets~\citep{maoposition,dong2024universal,liu2023towards}.
To explore this emerging paradigm, existing studies can be broadly categorized into two main lines of research: those that leverage LLMs as the backbone for GFMs, and those that rely on GNNs to construct GFMs.

\paragraph{LLM-based GFMs.}
Some methods investigate their feasibility by leveraging LLMs~\citep{zhu2024graphclip,tan2024musegraph,chen2024llaga,wang2024llms}. 
For example,
MuseGraph unifies graph descriptions, chain-of-thought generation, and graph-aware instruction tuning to enable LLMs on diverse graph mining tasks.
LLaGA introduces lightweight projectors to align graph embeddings with LLM representations.
GOFA~\citep{kong2024gofa} and ZeroG~\citep{li2024zerog} adapt language models through fine-tuning to enhance their transferability in zero-shot settings.
These approaches benefit from the powerful language understanding of LLMs, but they often struggle with structural complexity in graphs—particularly as graph size grows or sequence length increases.

\paragraph{GNN-based GFMs.}
Another line of work focuses on utilizing GNNs to build GFMs~\citep{zhao2024graphany}.
Some methods generalize pre-trained models across tasks via graph prompts
\citep{zhu2024relief,gong2024self,sun2023all,fang2024universal,liu2023graphprompt,liu2023one}. 
Others 
emphasize
the generalization capability across different datasets,
where various approaches utilize distinct graph elements as base units to learn how GFMs generalize to different datasets or domains.
For node-level generalization, 
Prodigy~\citep{huang2024prodigy} constructs task graphs that propagate label information from the support set to queries, thereby enabling in-context learning. 
OFA~\citep{liu2023one} introduces the nodes-of-interest (NOI) subgraph and the NOI prompt node, enabling joint training across multiple datasets and unifying different types of graph tasks.
OpenGraph~\citep{xia2024opengraph} develop a graph tokenizer that transforms input graphs into unified token sequences,
with each token representing a node and a semantic vector, thereby standardizing node distributions across different graphs.
For dataset-level generalization, OMOG~\citep{liu2024one} and AnyGraph~\citep{xia2024anygraph} both adopt Mixture-of-experts (MoE)~\citep{cai2024survey}, 
training a separate model for each pretraining graph to capture its unique structural and feature characteristics. 
GCOPE~\citep{zhao2024all} introduces coordinators to merge multiple graph datasets into a single large graph during pretraining, facilitating knowledge transfer across datasets. 
Moreover, 
MDGPT~\citep{yu2024text} employs domain tokens to align graph features across different domains and incorporates a dual prompt mechanism to further enhance adaptation to target tasks.

Despite these advances, most existing approaches rely on node-level or dataset-level pretraining,
which restricts flexibility and scalability.
In contrast, REEF introduces relation tokens as the fundamental unit for pre-training. This design allows REEF to unify diverse relations across datasets, while dataset-specific projectors and biases adapt to distributional differences. As a result, REEF not only supports comprehensive pre-training but also enables more effective fine-tuning and transfer, thereby enhancing generalization to new domains.

\section{Datasets}
\label{app:datasets}
Table~\ref{tab:datasets} summarizes the details.
Here is a detailed description of ten datasets in four domains:

\begin{itemize}[leftmargin=1.5em]
\item \textbf{Knowledge graphs:}
\begin{itemize}[leftmargin=1.5em]
\item FB15K237 is a knowledge graph containing relation triples and textual mentions of Freebase entity pairs. 
The raw text data for the nodes was collected from a GitHub repository\footnote{\url{https://github.com/villmow/datasets_knowledge_embedding/tree/master}}. 
The text feature of a node includes the entity name and description, while the text feature of an edge describes the relation type between entities.
\item WN18RR is a subset of WordNet, consisting of 11 relations and 40,943 entities. 
The raw text data for WN18RR nodes is also collected from the same GitHub repository\footnote{\url{https://github.com/villmow/datasets_knowledge_embedding/tree/master}}. 
Node and edge text features are processed in the same way as FB15K237.
\end{itemize}

\item \textbf{Citation networks:} Cora, Citeseer, and Pubmed are widely used citation network benchmarks. 
In these networks, nodes represent papers, and edges denote
citations of one paper by another. 
Node features are represented as bag-of-words vectors of the paper content, and node labels correspond to the paper’s research domain.

\begin{itemize}[leftmargin=1.5em]
\item 
Cora is a citation network in the computer science domain, where nodes represent papers and edges denote citation relationships. The dataset is categorized into seven research areas: case-based, genetic algorithms, neural networks, probabilistic methods, reinforcement learning, rule learning, and theory.
 \item 
 PubMed is a citation network of biomedical papers, categorized into three themes: diabetes, experimental diabetes, and type 1/type 2 diabetes. Edges indicate citation relationships between papers.
 \item CiteSeer is a citation network consisting of academic papers and their citation relationships in the computer science domain. 
 The dataset is categorized into six research fields: agents, artificial intelligence, database, information retrieval, machine learning, and human–computer interaction.
\end{itemize}

\item \textbf{WebKB\footnote{\url{http://www.cs.cmu.edu/afs/cs.cmu.edu/project/theo-11/www/wwkb}}:}
The WebKB dataset was collected by Carnegie Mellon University from computer science departments of various universities, consisting of three sub-datasets: Cornell, Texas, and Wisconsin. In these datasets, nodes represent web pages, and edges correspond to hyperlinks between them. The feature vectors of the nodes are derived from the bag-of-words representation of the web pages. These web pages are manually classified into five categories: student, project, course, staff, and faculty. 

\item \textbf{Amazon:} Amazon Computers and Amazon Photo are segments of the Amazon co-purchase graph, where nodes represent goods, edges indicate that two goods are frequently bought
together, node features are bag-of-words encoded product reviews, and class labels are given by the
product category.

\end{itemize}

\begin{table}[hbtp]
    \centering
    \caption{Statistics of datasets.}
    \label{tab:datasets}
    \resizebox{0.7\textwidth}{!}{
    \begin{tabular}{llllrrl}
        \toprule
Dataset & Domain & \#Nodes & \#Edges & \#Classes & Task & Type \\
\midrule
FB15K237 & Knowledge & 14,541 & 310,116 & 237 & Link & $\mathcal{G}_{\text{source}}$ \\
WN18RR & Knowledge & 40,943 & 93,003 & 11 & Link & $\mathcal{G}_{\text{source}}$ \\
PubMed & Citation & 19,717 & 44,338 & 3 & Node & $\mathcal{G}_{\text{source}}$ \\
Citeseer & Citation & 3,237 & 9,104  & 6 & Node & $\mathcal{G}_{\text{source}}$ \\
Cora & Citation & 2,708 & 10,556 & 7 & Node & $\mathcal{G}_{\text{target}}$ \\
Wisconsin & WebKB & 251 & 515 & 5 & Node & $\mathcal{G}_{\text{source}}$ \\
Texas & WebKB & 183 & 325 & 5 & Node & $\mathcal{G}_{\text{source}}$ \\
Cornell & WebKB & 183 & 298 & 5 & Node & $\mathcal{G}_{\text{target}}$ \\
Photo & Amazon & 7,650 & 238,162 & 8 & Node & $\mathcal{G}_{\text{source}}$ \\
Computers & Amazon & 13,752 & 491,722 & 10 & Node & $\mathcal{G}_{\text{target}}$ \\
        \bottomrule
    \end{tabular}
    }
\end{table}

\section{Baselines}
\label{app:baselines}
We compare our proposed framework with three groups of baseline models:
(1) General supervised GNN methods
including GCN~\citep{gcn}, GAT~\citep{gat}, and RGCN~\citep{rgcn}.
These models rely on task-specific supervision and serve as fundamental GNNs for graph representation learning.
(2) Graph contrastive learning methods like GraphCL~\citep{you2020graph} and SimGRACE~\citep{xia2022simgrace}, which leverage contrastive objectives to learn graph representations in a self-supervised manner.
The learned representations are then fine-tuned on downstream tasks to improve task-oriented performance.
(3) Graph pre-training methods for transfer: GCOPE~\citep{zhao2024all} employs a pretraining framework for contrastive learning that integrates information across multiple datasets through 
coordinators, 
enabling knowledge transfer to downstream tasks.
MDGPT~\citep{yu2024text} leverages domain tokens to align graph representations from different domains and adopts a dual-prompt strategy to further improve adaptation to the target task.

\section{Experimental Settings}
\label{app:exp_setting}
In our proposed framework~\model,
we implement it with Pytorch and adopt the Adam optimizer for training.
We fix the number of layers of the graph neural network $L$ at 2 for REEF and other baselines.
To unify feature representations across different datasets, we apply Singular Value Decomposition (SVD) to reduce the initial feature dimensions to 128. 
For REEF, we set the learning rate to 0.0002, the dropout rate to 0.5, the hidden dimension to 64,
batch size to 128,
and train for 100 epochs.
Each of the three hypernetworks ${\mathcal{F_{\text{Agg}}}, \mathcal{F_{\text{Cls}}}, \mathcal{F_{\text{Proj}}} }$ and the feature bias transformation function $\mathcal{F}_{\text{Feat}}$ adopts a two-layer MLP structure.
The edge mask rate in graph data augmentation is set to 0.2.
For fairness, we run all the experiments on a
server with 80G memory and a single Tesla A800 GPU.
For a detailed description of the relations, see Table~\ref{tab:relation_description}.

\model~is pre-trained on seven datasets: FB15K237, WN18RR, Pubmed, Citeseer, Wisconsin, Texas, and Photo, using only their training splits. 
FB15K237 and WN18RR follow their original train/validation/test splits (142k/47k/48k), while Wisconsin and Texas use the official dataset splits. Pubmed, Citeseer, and Photo are divided using a 6:2:2 ratio.

For the relation vocabulary, the relation tokens of the aggregators and classifiers for FB15K237 and WN18RR are derived from the edge descriptions in their original datasets.
As for Pubmed and Citeseer, 
they are based on the descriptions of ``citation'' and ``paper classification'', respectively.
Similarly,
for Wisconsin and Texas,
the relation token for the aggregator corresponds to the description of ``hyperlink'',
while the relation token for the classifier is based on the description of ``web page classification.''
In the case of Photo,
relation tokens are derived from the descriptions of ``co-purchase'' and ``product classification.''
The transfer learning datasets $\mathcal{G}_{\text{target}}$ include Cora, Cornell, and Computers, whose aggregators and classifiers are consistent with the dataset from the same domain.
In total, the size of the relation vocabulary in this pre-training setting is 254.

The relation types are not manually annotated on individual samples; they are determined by the relation schema and task semantics provided by each dataset.
For knowledge graphs, we directly use the original edge-relation types defined in FB15K237 and WN18RR.
For the remaining graph domains, we define relation types according to their dataset-level edge and prediction semantics, such as ``citation'' and ``paper classification,'' ``hyperlink'' and ``web page classification,'' and ``co-purchase'' and ``product classification.''
For each relation type, we prompt an LLM to generate a natural-language description based on the corresponding dataset background, the official relation name, and the task semantics; these descriptions are then encoded by Sentence-BERT and used as the inputs to \model's hypernetworks.
Concretely, the 254 pre-training relation tokens comprise 237 relations from FB15K237, 11 from WN18RR, and six domain-level relation tokens covering the aggregation and classification semantics of the citation, webpage, and Amazon co-purchase domains.

For transfer learning,
we follow the GCOPE setup, adopting 
1-shot learning on the target datasets $\mathcal{G}_{\text{target}}$,
splitting the remaining data into validation and test sets with a 1:9 ratio. 
The transfer learning target datasets include Cora, Cornell, and Computers, whose aggregators and classifiers share the same domain-level relation semantics as the corresponding pre-training datasets.

For the transfer learning baselines, 
we follow the original experimental settings of GCOPE~\citep{zhao2024all} and MDGPT~\citep{yu2024text}, respectively.
Specifically, GCOPE involves ten datasets; following its protocol, we pretrain the model on nine datasets and conduct transfer evaluation on the remaining one.
Similarly, MDGPT involves seven datasets, where the model is pretrained on six datasets and evaluated on the remaining one.
Since the Cornell dataset is not included among the seven datasets used in MDGPT, we pretrain on these seven datasets and evaluate the transfer performance on Cornell.

\section{Supplementary Experiments}
\label{app:supp_exp}
As a complement to Section~\ref{sec:exp}, we provide additional experimental results and analyses in this section.

\begin{itemize}[leftmargin=1.5em]
    \item \textbf{RQ7:} How does REEF perform on downstream link prediction tasks?
    \item \textbf{RQ8:} How do scaling laws and dataset diversity affect the performance of REEF during pre-training?
    \item \textbf{RQ9:} How does the REEF framework generalize when encountering out-of-domain relations?
    \item \textbf{RQ10:} What is the impact of hidden dimensions on the performance of the framework?
    \item \textbf{RQ11:} What relational representations are learned by REEF?
    \item \textbf{RQ12:} What is the computational complexity and scalability of REEF?
     \item \textbf{RQ13:} How does the performance vary across differenet datasets perform during pre-training?
    \item \textbf{RQ14:} How does the hypernetwork-based aggregator compare with simpler relation-conditioned alternatives?
    \item \textbf{RQ15:} How do nonlinear activations in the generated message-passing layers affect performance?
    \item \textbf{RQ16:} What are the empirical trainable parameters, runtime, and memory cost of REEF?
    \item \textbf{RQ17:} How does REEF compare with GNNs specifically designed for heterophilous graphs?
    \item \textbf{RQ18:} Can REEF be extended to heterogeneous graphs with multiple node and edge types?
\end{itemize}

\begin{table}[hbtp]
\centering
\caption{Accuracy results on the link prediction task.}
\label{tab:pretrain_lp}
    \resizebox{1.0\textwidth}{!}{
\begin{tabular}{l|ccccccccc|c}
\toprule
Dataset & Linear & GCN & GAT & GIN & DGI & BGRL & GraphMAE & GIANT & GFT & \textbf{\model} \\
\midrule
FB15K237 & 87.39 & 82.22 & 88.93 & 83.21 & 81.34 & 80.66 & 85.30 & 87.45 & 89.72 & \textbf{91.04} \\
WN18RR & 78.50 & 73.79 & 80.16 & 74.02 & 75.75 & 75.44 & 78.99 & 84.36 & 91.91 & \textbf{94.70} \\
\bottomrule
\end{tabular}
}
\end{table}

\subsection{Link Prediction (RQ7)}
\label{app:lp}
We evaluate the performance of~\model~on the downstream task of link prediction. 
Relevant baseline models are categorized into three groups:
(1) Supervised GNNs and MLP: This group includes a basic MLP (Linear), GCN~\citep{gcn}, GAT~\citep{gat}, and GIN~\citep{gin}.
(2) Self-supervised GNNs: 
DGI~\citep{dgi} leverages contrastive learning between global graph summaries and local node patches. 
BGRL~\citep{bgrl} adopts bootstrapping to predict representations of the same node across different augmented views. GraphMAE~\citep{hou2022graphmae} reconstructs node features based on structural information. 
GIANT~\citep{giant} integrates pretrained language models with GNNs under a self-supervised learning paradigm.
(3) Graph Foundation Model:
GFT~\citep{gft} builds a GFM by treating computation trees as transferable vocabulary units to promote generalization and mitigate negative transfer.
Here, GCOPE~\citep{zhao2024all} and
MDGPT~\citep{yu2024text} are not used because their original papers do not include link prediction experiments.
For fairness, 
the baseline results are reported from~\citep{gft}, and REEF is evaluated under the same data split setup.

Table~\ref{tab:pretrain_lp} presents the accuracy results of various models evaluated on the link prediction task across two benchmark datasets, FB15K237 and WN18RR.
Among supervised and self-supervised baselines, performance varies moderately; for instance,
GFT achieves strong results with 89.72\% accuracy on FB15K237 and 91.91\% on WN18RR. 
Building upon this foundation, 
REEF achieves the best performance across both datasets, with 91.04\% on FB15K237 and 94.70\% on WN18RR, significantly outperforming all baselines.
These results highlight REEF’s superior generalization ability and its effectiveness in link prediction tasks across diverse relational graphs.

\subsection{Scaling Laws in Performance Across Different Dataset Scales (RQ8)}

\label{app:scaling}

\begin{figure}
  \centering
  \includegraphics[width=0.7\linewidth]{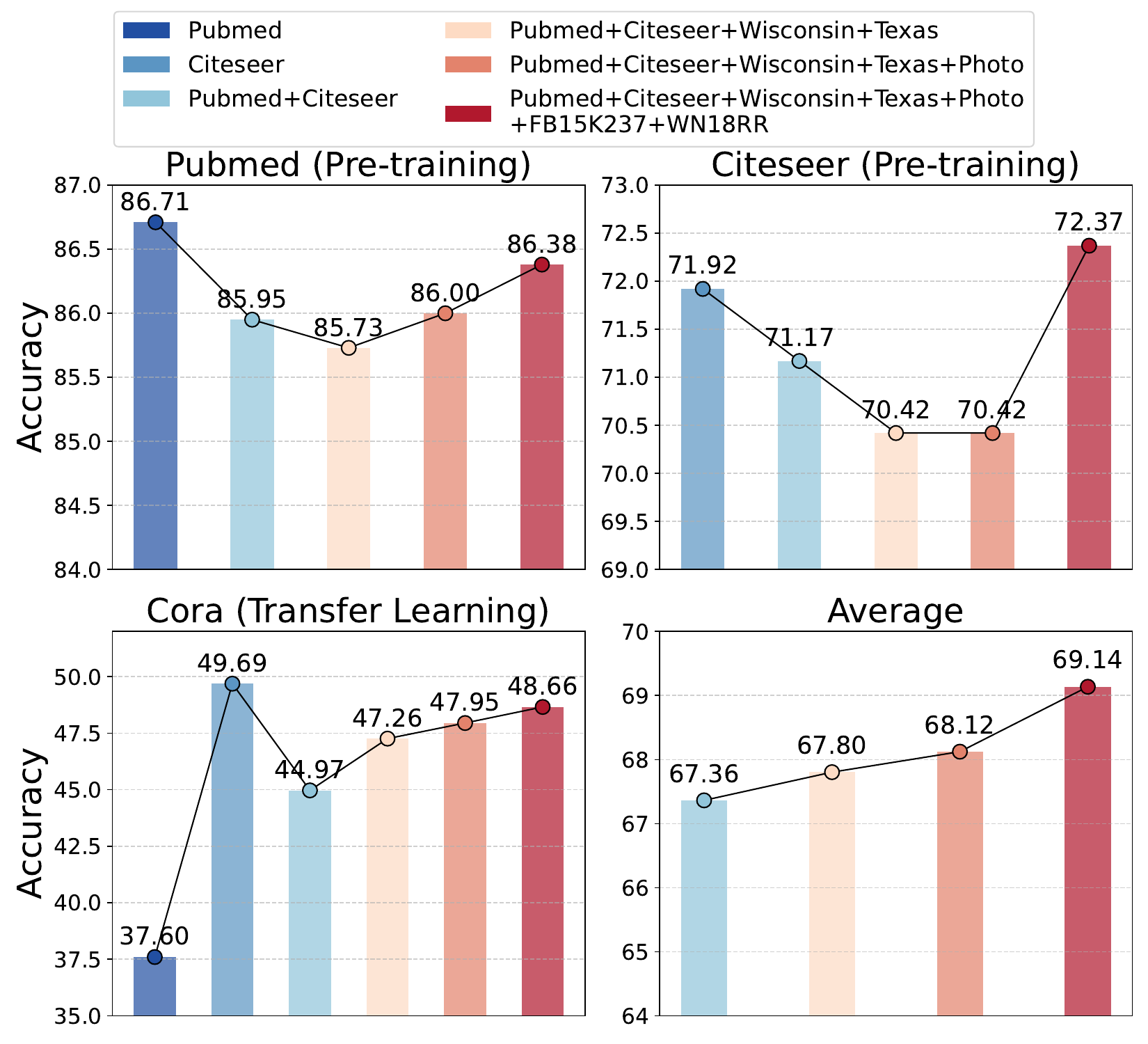}
  \captionof{figure}{Performance comparison across different dataset scales in the citation domain.}
  \label{fig:dataset_scale}
\end{figure}

We examine REEF under six pre-training settings covering different dataset scales (see Figure~\ref{fig:dataset_scale}), and evaluate pre-training on Pubmed and Citeseer, transfer learning on Cora, and average performance across datasets.
The key observations are summarized as follows:
(1) \textbf{Pre-training performace:} Pubmed and Citeseer, when used individually as pre-training datasets, both exhibit strong downstream performance.
As more datasets are added, performance may fluctuate due to distributional shifts in multi-dataset training, but it consistently improves as the dataset scale grows.
This rebound stems from an enriched vocabulary and exposure to diverse relations, which help the model learn more generalizable representations.
(2) \textbf{Transfer learning performance:}
Pre-training on Citeseer alone achieves the best transfer accuracy on Cora, likely due to topological and semantic similarities.
In contrast, although Pubmed is also a citation network, 
it shows the worst transfer performance. 
This is likely because Citeseer and Cora both originate from the computer science domain, whereas Pubmed comes from the medical domain and differs more substantially from Cora.
Notably,
as the training dataset scale increases, 
the performance of Cora keeps improving.
This phenomenon shows that incorporating training data with diverse relations can enhance the model’s transferability, leading to better cross-dataset generalization.
(3)	\textbf{Overall average performance:} 
The average accuracy improves consistently as relational coverage expands, revealing a clear relation-driven scaling behavior. 
While minor performance drops may occur on specific tasks, such fluctuations are acceptable in the foundation model setting, which prioritizes broad generalization and overall performance across diverse tasks. 
These results demonstrate that enriching relational diversity systematically enhances the generalizability and transferability of REEF, reflecting a clear scaling trend in overall foundation model performance.

\begin{table}[ht]
\caption{Transfer Learning Performance on Out-of-Domain Relations for Cornell and Computers Datasets.}
\label{tab:transfer_outdomain}
\centering
\resizebox{1.0\textwidth}{!}{
\begin{tabular}{l|c|c|c|c|c|c}
\toprule
\multirow{2}{*}{Pre-training datasets} 
& \multicolumn{3}{c|}{\textbf{Cornell (\#Labels = 5)}} & \multicolumn{3}{c}{\textbf{Computers (\#Labels = 10)}} \\
 \cmidrule(lr){2-4} \cmidrule(lr){5-7} 
 & Acc & AUC & F1 & Acc & AUC & F1 \\
\hline
Pubmed+Citeseer+Wisconsin+Texas+ & 
\multirow{2}{*}{48.65$\pm$7.76} & \multirow{2}{*}{70.16$\pm$2.82} & \multirow{2}{*}{36.49$\pm$6.77} & \multirow{2}{*}{47.72$\pm$5.61} & \multirow{2}{*}{84.80$\pm$3.22} & \multirow{2}{*}{44.03$\pm$2.49} \\
Photo+FB15K237+WN18RR (REEF) & & & & & & \\ 
\hline
Pubmed+Citeseer+FB15K237+WN18RR & 34.58$\pm$12.87 & 57.45$\pm$4.57 & 20.29$\pm$6.34 & 26.40$\pm$11.19 & 47.65$\pm$6.20 & 6.10$\pm$1.35 \\

\bottomrule
\end{tabular}
}
\end{table}

\subsection{Transfer Performance on Out-of-Domain Relations (RQ9)}
\label{app:transfer_outdomain}

In this experiment, we perform transfer learning on out-of-domain relationships on Cornell and Computers. 
During fine-tuning, we separately introduce the relations corresponding to the Cornell dataset, including ``hyperlinks'' and ``webpage classification'', and those from the Computers dataset, including ``co-purchase'' and ``product classification.''
The pre-training datasets include multiple datasets,
including 
Pubmed, Citeseer, FB15K237, and WN18RR,
which include both citation and knowledge graph domains.
As shown in Table~\ref{tab:transfer_outdomain}, we observe the following:

(1)	Effectiveness of Fine-tuning and Generalization:
Since Cornell has 5 labels and Computers has 10, the results show that fine-tuning is effective across these datasets.
This demonstrates that even when the pre-training data does not contain relationships specifically corresponding to the out-of-domain relations in the target dataset, our proposed framework~\model~can still effectively transfer knowledge and adapt to new, unseen relations, exhibiting strong generalization ability. 

(2) Importance of Relations in the Transfer Dataset:
The effectiveness of transfer learning is closely tied to the quality and diversity of the relationship data in the pre-training dataset. The more diverse the relations in the pre-training dataset, the better the model performs in transfer learning tasks. 
By comparing different pre-training configurations, we find that pre-training with a dataset containing a wider variety of relations leads to significantly better performance on the target tasks. 
For instance, the pre-training configuration involving Pubmed, Citeseer, Wisconsin, Texas, FB15K237, and WN18RR outperforms another configuration with less pre-training datasets in three metrics, demonstrating the importance of rich relational data in pre-training.

These experimental results indicate that the proposed framework~\model~has strong adaptability when handling transfer learning tasks, particularly when faced with new or unseen relationships, allowing it to achieve good performance transfer. 
Additionally, the diversity of relationships in the pre-training dataset plays a crucial role in enhancing the model’s performance.

\begin{figure}[hbtp]
    \centering
\includegraphics[width=0.75\linewidth]{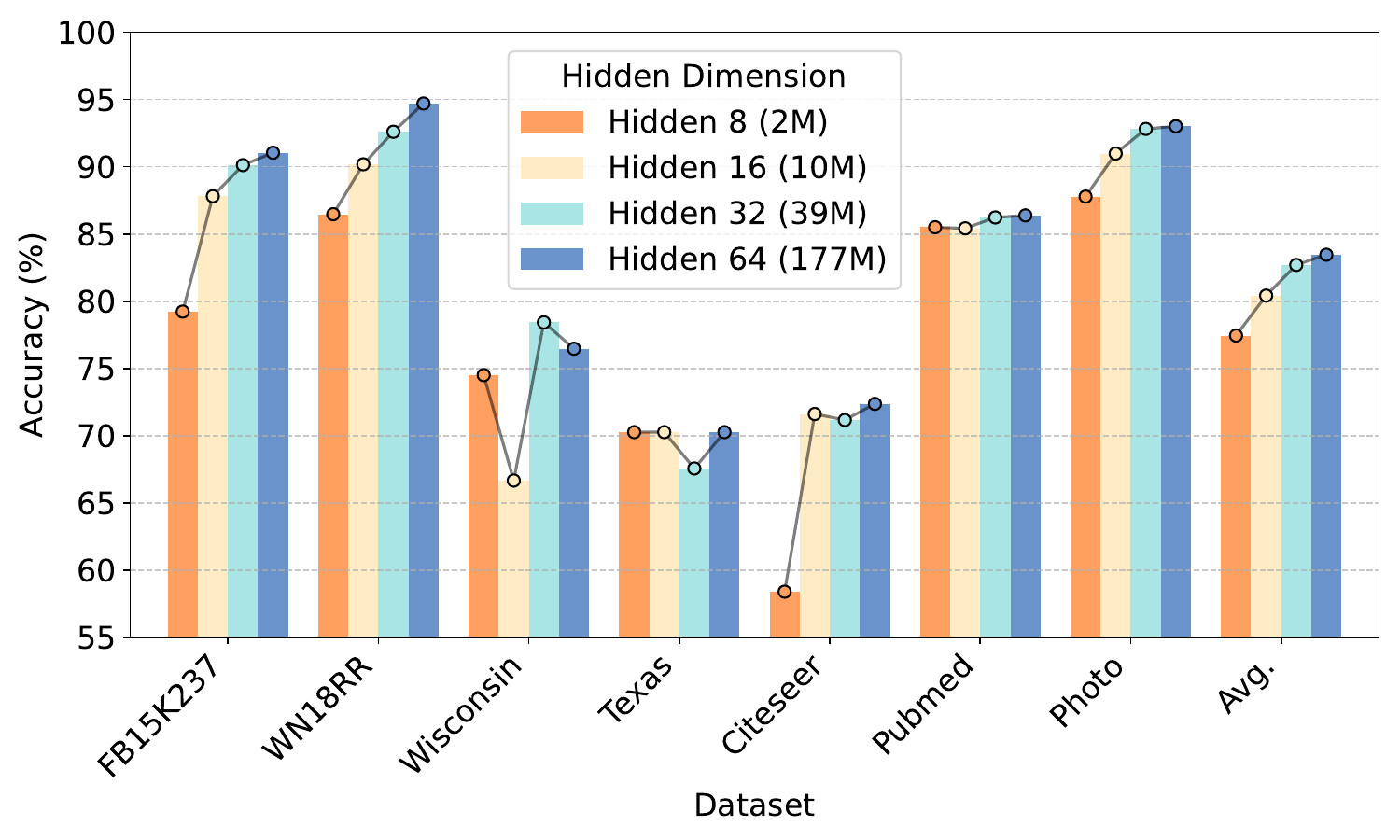}
    \caption{Performance of different datasets with different hidden dimensions. Model size is indicated in parentheses.}
    \label{fig:hidden}
\end{figure}

\subsection{Impact of Hidden Dimensions on Model's performance (RQ10)}
\label{app:hyper_hidden}

We set the hidden dimension in our proposed framework~\model~to 8, 16, 32, and 64 to compare the model’s performance. 
In Figure~\ref{fig:hidden},
we can see that as the hidden dimension increases, most datasets exhibit an upward trend in accuracy, and the overall average performance (Avg) also improves.
This indicates a scaling law in REEF, where larger model capacity and increased parameter size generally lead to better overall performance.
For smaller datasets such as Wisconsin and Texas, model performance is more sensitive to the hidden dimension, resulting in greater fluctuations. 
This instability may stem from the high variance caused by the small dataset size, leading to differences in generalization ability across different hidden dimensions.

\subsection{Visualization of Representations (RQ11)}
\label{app:visual}

\begin{figure}[hbtp]
  \centering
  \includegraphics[width=0.75\linewidth]{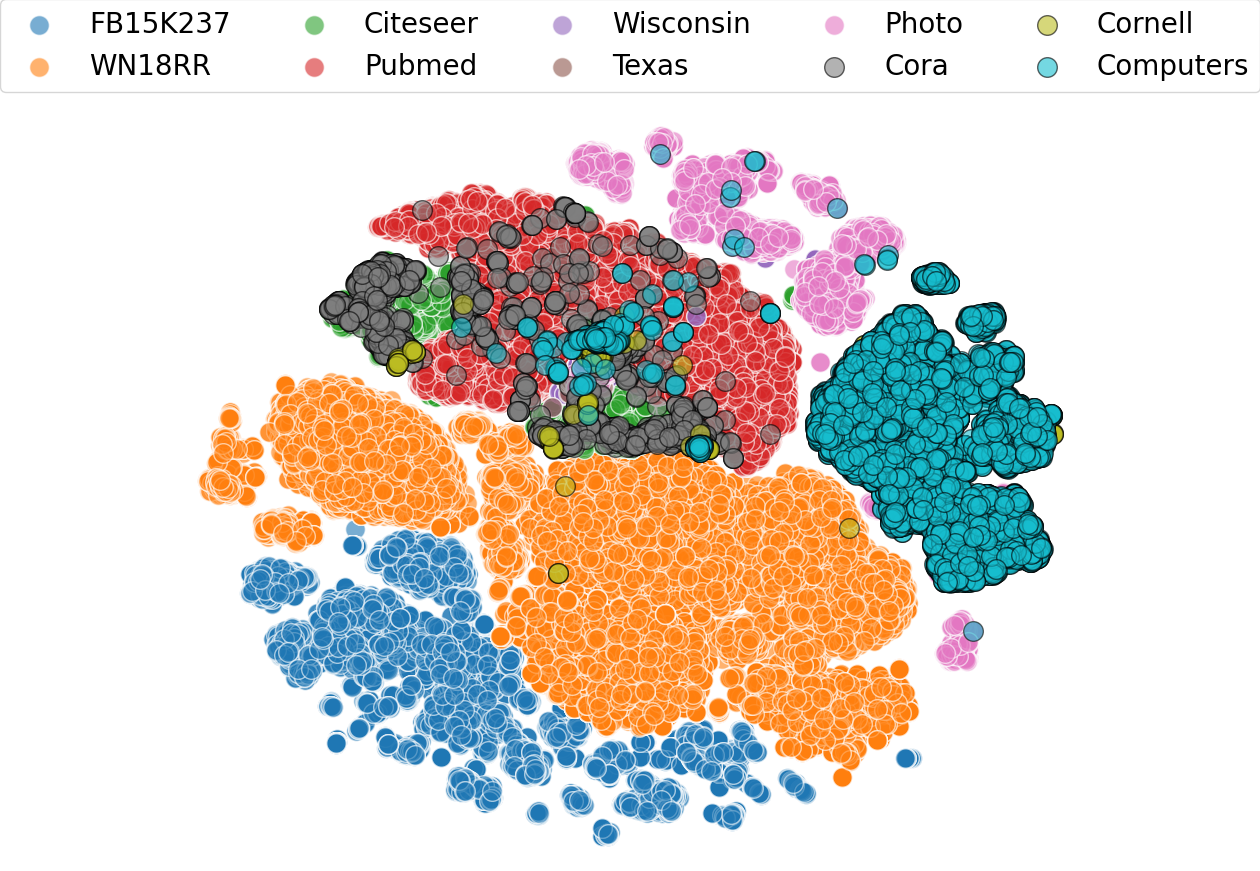}
  \caption{T-SNE visualization of representations after pretraining.
}
  \label{fig:tsne}
\end{figure}

The t-SNE visualizations in Figure~\ref{fig:tsne} illustrate the distribution of representations across different datasets after pre-training. 
We can see that:
(1)
The pretraining datasets ($\mathcal{G}_{\text{source}}$) form well-structured and distinct clusters.
Datasets within the same domain are positioned more closely, while datasets from different domains are farther apart,
demonstrating the model’s ability to differentiate between different types of data,
which is crucial for generalization.
Moreover, different relations exhibit a certain degree of separation, with knowledge graph datasets clustering closely together and citation datasets forming distinct groups.
(2)
Before fine-tuning, the target datasets ($\mathcal{G}_{\text{target}}$) are projected into the pretraining space.
Notably, the representations of the Cora and Computers datasets exhibit strong internal clustering and align closely with their corresponding source-domain clusters, reflecting consistency in the learned relational structures.
These findings suggest that the pre-trained model effectively captures meaningful relational patterns and exhibits strong feature transferability, providing a solid foundation for downstream tasks.

\subsection{Complexity Analysis (RQ12)}
\label{app:complex}
For our proposed framework~\model,
the parameters are $\{ \Theta_{r,\text{Agg}}, \allowbreak
\Theta_{r,\text{Cls}}, \Theta_{\mathcal{G},\text{Proj}},
\Theta_{\text{Feat}}
\}$,
corresponding to three hypernetworks 
$\{
\mathcal{F}_{\text{Agg}}, \allowbreak
\mathcal{F}_{\text{Cls}},
\mathcal{F}_{\text{Proj}}
\}
$
and the 
feature bias transformation function
$
\mathcal{F}_{\text{Feat}}
$.
Each of these four functions is implemented as a two-layer MLP,
The computational complexity of one such function is
$
O(d^3)
$,
where $d$ represents the hidden dimension.
Thus, the total complexity contributed by these functions is:
$
O(
L\cdot|\mathcal{R_{{\text{source}}}}|\cdot d^3+|\mathcal{R_{{\text{source}}}}|\cdot d^3+L\cdot |\mathcal{G_{{\text{source}}}}|\cdot d^3+|\mathcal{G_{{\text{source}}}}|\cdot d^2)
$,
which simplifies to
$
O(
L\cdot(|\mathcal{R_{{\text{source}}}}| + |\mathcal{G_{{\text{source}}}}|)\cdot d^3)
$.
Here, 
$L$ denotes the number of layers in the GNN.
$|\mathcal{R_{{\text{source}}}}|$ and $|\mathcal{G_{{\text{source}}}}|$ represent the size of relation vocabulary and the number of pre-training datasets,
respectively.
Then,
for GNN components, the complexity is 
$
O(L \cdot(|\mathcal{E}_{\mathcal{G}_{\text{source}}}|\cdot d + |\mathcal{R_{{\text{source}}}}|\cdot |\mathcal{V}_{\mathcal{G}_{\text{source}}}|\cdot d^2)),
$
where $|\mathcal{E}_{\mathcal{G}_{\text{source}}}|$ denote the total number of edges, and $|\mathcal{V}_{\mathcal{G}_{\text{source}}}|$ represents the total number of nodes in the pre-training datasets.
Combining these, the overall complexity of the framework is
$
O\Big(
L\cdot
\Big(
(|\mathcal{R_{{\text{source}}}}| + |\mathcal{G_{{\text{source}}}}|)\cdot d^3
+
|\mathcal{E}_{\mathcal{G}_{\text{source}}}|\cdot d 
+ |\mathcal{R_{{\text{source}}}}|\cdot |\mathcal{V}_{\mathcal{G}_{\text{source}}}|\cdot d^2
\Big)
\Big).
$

In summary, compared to RGCN, the time complexity of our proposed framework includes two additional components: the relation vocabulary and the feature enhancement modules, 
which are related to the size of the relation vocabulary and the number of datasets. 
The main factor influencing complexity remains the data scale, 
including the number of nodes and edges in the pre-training datasets.
Overall, our framework achieves a favorable complexity, making it well-suited for graph foundation models and scalable for large-scale graph data training.

\begin{figure}[hbtp]
    \centering
\includegraphics[width=0.80\linewidth]{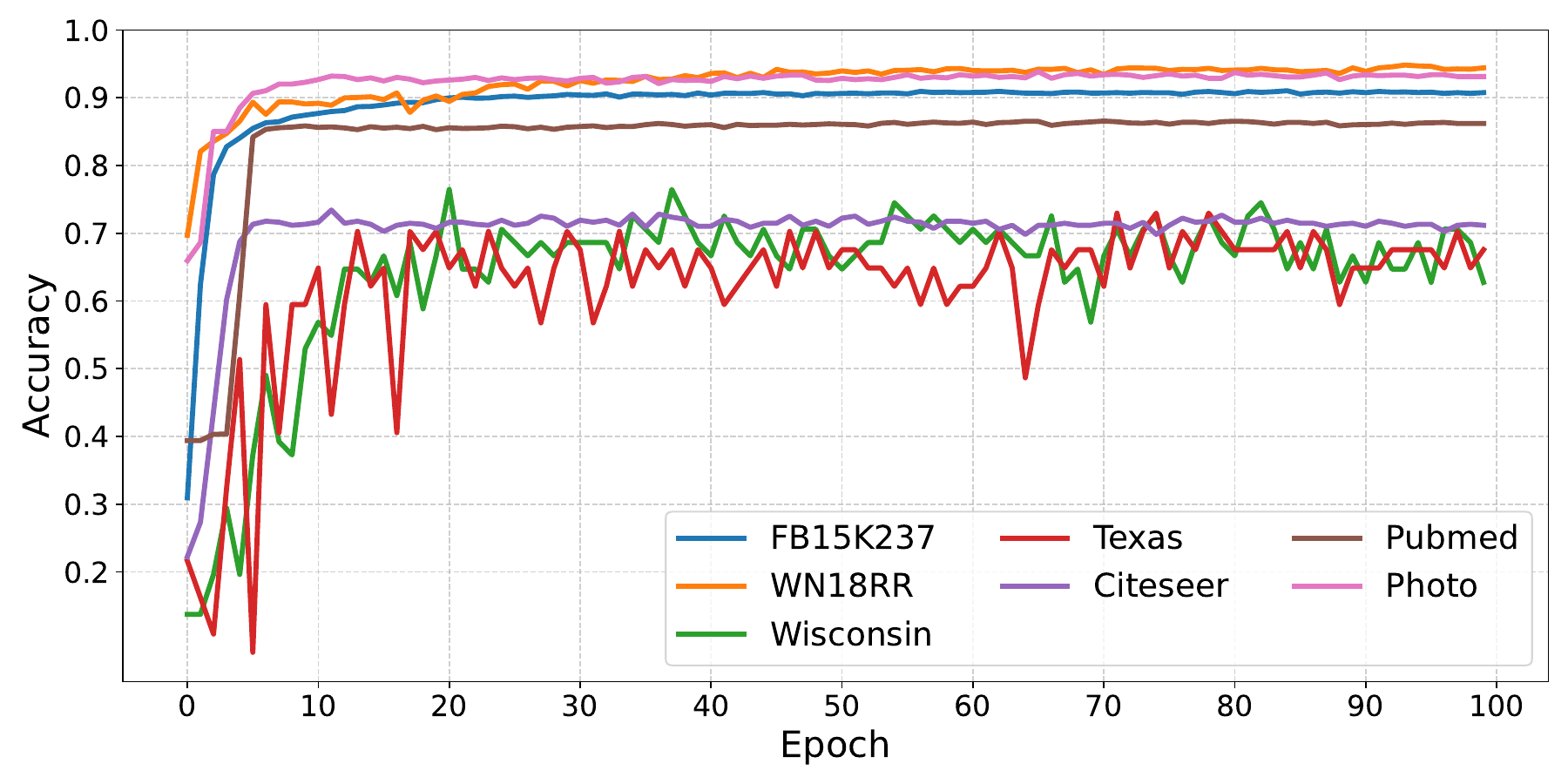}
    \caption{Test accuracy of different datasets during pre-training.}
    \label{fig:train_testacc}
\end{figure}

\subsection{Performance Trends across Different Datasets During Pre-training (RQ13)}
\label{app:trends}
Figure~\ref{fig:train_testacc} presents the test accuracy trends of the framework~\model~ across multiple datasets during the pre-training process. 
Overall, the test accuracy of all datasets increases rapidly within the first 10 epochs and then stabilizes, indicating that the model can quickly learn effective features in the early training stages. 
Among them, FB15K237, WN18RR, Photo, and Pubmed achieve high accuracy early and maintain stability throughout training, demonstrating good convergence properties. 
Notably, 
the Photo dataset stabilizes around 0.9,
likely due to the relatively simple nature of its task and the large size of its training data.
In contrast, the test accuracy of the Wisconsin and Texas datasets exhibits significant fluctuations, with final accuracy around 0.6-0.7, indicating weaker generalization ability on these datasets, potentially due to their small scale. 
Additionally, the Citeseer dataset stabilizes around 0.7, showing more stability than Wisconsin and Texas but still underperforming compared to FB15K237 and WN18RR. 
Regarding convergence stability, the error curves of Photo, Pubmed, and WN18RR are relatively smooth during training, suggesting that the model is more stable on these datasets, likely due to their structured data distributions or the model’s strong learning capabilities for these types of data. 
In contrast, Wisconsin and Texas show greater fluctuations, especially Texas, indicating less stable training on these datasets, possibly due to their small size or complex distributions, making the model prone to overfitting or sensitive to random variations. 

Overall, the pre-trained model exhibits significant performance differences across datasets, excelling on structured knowledge graphs (FB15K237, WN18RR) and certain domain-specific datasets (Photo, Pubmed) while being less stable on small-scale datasets such as Texas and Wisconsin.

\begin{table}[hbtp]
\centering
\caption{Comparison of different relation-conditioned aggregator designs under the same pre-training and transfer protocol. 
Only the aggregator is replaced; the classifier hypernetwork, dataset projector, and feature bias remain unchanged. 
``Pre.'' refers to the pretraining average, ``Trans.'' to the transfer average, and ``All'' to the overall average. The best score is highlighted in bold
.}
\label{tab:agg_compare}
\begin{tabular}{l|ccc}
\toprule
\textbf{Aggregator} & \textbf{Pre.} & \textbf{Trans.} & \textbf{All} \\
\midrule
Shared             & 67.46 & 29.80 & 56.16 \\
Concat             & 80.60 & 39.39 & 68.23 \\
Relation-ID        & 83.24 & 33.51 & 68.32 \\
Cross-Attention    & 65.10 & 34.29 & 55.86 \\
\midrule
\model~(Hypernetwork) & \textbf{83.46} & \textbf{48.34} & \textbf{72.93} \\
\bottomrule
\end{tabular}
\end{table}

\subsection{Comparison of Aggregator Designs (RQ14)}
\label{app:agg_compare}
A natural question is whether \model's transfer performance depends on its specific hypernetwork-based aggregator, or whether simpler relation-conditioned alternatives would suffice.
To isolate this factor, we conduct a controlled comparison in which \emph{only} the aggregator is replaced, while the classifier hypernetwork, dataset projector, feature bias, and all training and transfer protocols remain unchanged.
We consider four representative alternatives:
(1) \textbf{Shared}: a single transformation matrix shared across all relations ($\mathbf{W}_{\text{shared}} \mathbf{h}_j$);
(2) \textbf{Concat}: an MLP over the concatenation of the neighbor representation and the projected relation embedding ($\text{MLP}([\mathbf{h}_j \| \mathbf{P}\mathbf{z}_r])$);
(3) \textbf{Relation-ID}: independently learned relation-specific transformations ($\mathbf{W}_r \mathbf{h}_j$) without semantic embeddings;
(4) \textbf{Cross-Attention}: the target-node representation is used as the query and the relation-aware neighbor representations as keys/values to compute relation-conditioned attention, instead of relation-specific transformation matrices.

As shown in Table~\ref{tab:agg_compare}, neither increasing relation-specific capacity (Relation-ID) nor simply conditioning message computation on relation embeddings (Concat and Cross-Attention) matches the performance of \model.
Instead, \model~consistently achieves the best pretraining, transfer, and overall averages.
The gap is most pronounced on transfer (48.34 vs.\ at most 39.39 for the alternatives), indicating that \emph{generating} relation-specific aggregation parameters from semantic relation embeddings, rather than sharing, indexing, or attending over them, is the key to \model's superior transferability.

\begin{table*}[hbtp]
\centering
\caption{Effect of the activation function used in the generated message-passing layers. All variants share the same pre-training protocol and differ only in the activation applied within the aggregator. The best average is highlighted in bold.}
\label{tab:activation}
\resizebox{1.\textwidth}{!}{
\begin{tabular}{l|ccccccc|c}
\toprule
\textbf{Activation} & \textbf{FB15K237} & \textbf{WN18RR} & \textbf{Wisconsin} & \textbf{Texas} & \textbf{Citeseer} & \textbf{Pubmed} & \textbf{Photo} & \textbf{Avg.} \\
\midrule
Linear & 91.04 & 94.70 & 76.47 & 70.27 & 72.37 & 86.38 & 93.01 & 83.46 \\
ReLU   & 90.47 & 92.38 & 72.55 & 67.57 & 78.10 & 89.00 & 94.34 & 83.49 \\
GELU   & 90.80 & 93.43 & 70.59 & 75.68 & 74.20 & 88.50 & 94.66 & \textbf{83.98} \\
\bottomrule
\end{tabular}
}
\end{table*}

\subsection{Effect of Nonlinear Activations (RQ15)}
\label{app:activation}
\model~is not restricted to linear message passing.
The submitted model adopts a lightweight linear instantiation, but nonlinear transformations can be incorporated directly into the generated message-passing layers.
Under the same pre-training protocol, we compare linear propagation with ReLU and GELU activations applied within the generated aggregator.
As shown in Table~\ref{tab:activation}, the three variants achieve comparable overall performance (83.46 / 83.49 / 83.98), with no activation consistently dominating across datasets.
These results show that \model~remains effective with both linear and nonlinear GNN predictors, confirming that linear propagation is an implementation choice rather than a methodological restriction.

\begin{table*}[hbtp]
\centering
\caption{Trainable parameters and total training resources of \model~compared with a representative GNN-based (GCOPE) and an LLM-based (GOFA) graph foundation model. GOFA numbers are as reported by its authors.}
\label{tab:efficiency}
\resizebox{1.\textwidth}{!}{
\begin{tabular}{l|l|l|l}
\toprule
\textbf{Method} & \textbf{Paradigm} & \textbf{Trainable params} & \textbf{Training resources} \\
\midrule
GCOPE~\citep{zhao2024all} & GNN-based & 93.3K & 34 min, 1$\times$A800 80GB; 23.40 MiB peak \\
GOFA~\citep{kong2024gofa} & LLM-based & 1.01B pretrain / 852.7M instruction tuning & Pre-training: 4 days, 4$\times$A100 80GB \\
\model & Hypernetwork + GNN-based & 177.4M & 8 h, 1$\times$A800 80GB; 48.21 GiB peak \\
\bottomrule
\end{tabular}
}
\end{table*}

\subsection{Empirical Efficiency Analysis (RQ16)}
\label{app:efficiency}
To complement the asymptotic analysis in Appendix~\ref{app:complex}, we report the empirical trainable parameters, wall-clock training time, and peak GPU memory of \model, and contextualize them against a representative GNN-based model (GCOPE) and an LLM-based graph foundation model (GOFA).
As summarized in Table~\ref{tab:efficiency}, \model~introduces more trainable parameters than a fixed GNN backbone such as GCOPE, because it uses semantic hypernetworks to generate relation- and dataset-specific GNN parameters, which is what enables its stronger cross-dataset and cross-domain transfer.
At the same time, \model~avoids the dependence on a multi-billion-parameter autoregressive LLM and the costly multi-GPU pre-training of LLM-based models such as GOFA, while achieving stronger overall transfer performance.
\model~therefore bridges conventional GNN-based pre-training and LLM-based graph foundation models: it converts its additional modeling capacity into effective transferability while requiring substantially fewer training resources than LLM-based counterparts.

\begin{table}[hbtp]
\centering
\caption{Node classification accuracy on eight heterophilous datasets, comparing \model~with GNNs specifically designed for heterophily. The best score in each row is highlighted in bold.}
\label{tab:heterophily}
\resizebox{0.88\columnwidth}{!}{
\begin{tabular}{l|ccccc|c}
\toprule
\textbf{Dataset} & \textbf{H2GCN} & \textbf{FAGCN} & \textbf{GPR-GNN} & \textbf{LINKX} & \textbf{ACM-GCN} & \textbf{\model} \\
\midrule
Cornell      & 72.97 & 45.95 & 51.35 & \textbf{78.38} & 67.57 & \textbf{78.38} \\
Texas        & 78.38 & 64.86 & 59.46 & 75.68 & 78.38 & \textbf{81.08} \\
Wisconsin    & 82.35 & 66.67 & 70.59 & 62.75 & 70.59 & \textbf{84.31} \\
Chameleon    & 57.46 & 62.06 & 61.18 & \textbf{72.59} & 60.75 & 56.80 \\
Actor        & \textbf{37.96} & 36.32 & 34.74 & 29.67 & 37.24 & 33.82 \\
BlogCatalog  & 80.65 & 86.52 & \textbf{89.32} & 60.91 & 74.98 & 76.93 \\
RomanEmpire  & \textbf{77.18} & 47.90 & 69.27 & 33.41 & 73.72 & 75.68 \\
Ratings      & 44.34 & 36.80 & 44.08 & \textbf{51.77} & 36.80 & 41.47 \\
\midrule
Avg.         & \textbf{66.41} & 55.88 & 60.00 & 58.14 & 62.50 & 66.06 \\
\bottomrule
\end{tabular}
}
\end{table}

\subsection{Comparison with Heterophily-Specific GNNs (RQ17)}
\label{app:heterophily}
\model~performs strongly on heterophilous graphs even though it is not explicitly designed for this setting.
To examine this observation in greater detail, we compare \model~with five GNNs specifically designed for heterophily, i.e.,\ H2GCN~\citep{h2gcn}, FAGCN~\citep{fagcn}, GPR-GNN~\citep{gprgnn}, LINKX~\citep{linkx}, and ACM-GCN~\citep{acmgcn}, across eight heterophilous datasets.
As shown in Table~\ref{tab:heterophily}, \model~remains highly competitive, attaining an average accuracy of $66.06\%$, on par with the strongest heterophily-specific baseline (H2GCN, $66.41\%$) and clearly ahead of the others.
We attribute this to \model's relation-conditioned aggregation: rather than applying a uniform propagation rule, \model~generates context-dependent transformations from relation semantics, allowing different relational patterns to be modeled with different message-passing parameters.
Such adaptivity is particularly beneficial when neighboring nodes may carry dissimilar labels and uniform neighborhood aggregation becomes unreliable.

\begin{table}[hbtp]
\centering
\caption{Node classification performance of \model~on standard heterogeneous graph benchmarks (ACM and DBLP) under the original HGB splits. Each canonical edge type is treated as an individual relation token, and node-type-specific projections map heterogeneous node features into the shared input space.}
\label{tab:hetero_graph}
\begin{tabular}{l|ccc}
\toprule
\textbf{Dataset} & \textbf{Micro-F1} & \textbf{Macro-F1} & \textbf{AUROC} \\
\midrule
ACM  & 84.94 & 85.03 & 94.46 \\
DBLP & 81.09 & 80.46 & 93.17 \\
\bottomrule
\end{tabular}
\end{table}

\subsection{Extension to Heterogeneous Graphs (RQ18)}
\label{app:hetero_graph}
Although the main experiments focus on homogeneous graphs and knowledge graphs, \model's relation-centric formulation extends naturally to heterogeneous graphs.
We adopt a minimal heterogeneous adaptation: node features from different node types are first projected into the shared input space using node-type-specific projections, and each canonical edge type $r = (\tau(u), \rho, \tau(v))$ is treated as an individual relation with its own relation token.
The subsequent relation-conditioned message passing remains unchanged, preserving \model's semantic-to-parameter generation mechanism.
We evaluate this adaptation on two standard heterogeneous benchmarks, ACM and DBLP, under the original HGB splits.
As shown in Table~\ref{tab:hetero_graph}, \model~obtains $84.94\%$ Micro-F1, $85.03\%$ Macro-F1, and $94.46\%$ AUROC on ACM, and $81.09\%$ Micro-F1, $80.46\%$ Macro-F1, and $93.17\%$ AUROC on DBLP.
These results provide direct empirical evidence that relation tokens are applicable not only to homogeneous graphs and knowledge graphs, but also to heterogeneous graphs with multiple node types and canonical edge types.

\begin{table*}[t]
\centering
\caption{Performance comparison across 26 common datasets. Best results are highlighted in bold.}
\resizebox{0.65\textwidth}{!}{
\begin{tabular}{lcccc}
\toprule
Dataset & GAT & GraphAny & TS-Mean & REEF \\
\midrule
Actor        & 32.59 & 29.51 & 28.09 & \textbf{33.82} \\
AirBrazil    & 35.38 & 36.15 & 39.23 & \textbf{44.44} \\
AirEU        & 39.00 & 41.13 & 35.88 & \textbf{41.87} \\
AirUS        & 43.03 & \textbf{43.86} & 42.34 & 41.62 \\
AmzComp      & 70.94 & \textbf{82.00} & 81.37 & 77.71 \\
AmzPhoto     & 80.78 & 90.18 & 90.18 & \textbf{93.83} \\
AmzRatings   & 40.63 & \textbf{42.57} & 42.27 & 41.47 \\
ogbn-arxiv   & \textbf{57.93} & 57.79 & 56.33 & 50.42 \\
BlogCatalog  & \textbf{78.20} & 77.69 & 76.30 & 76.93 \\
Chameleon    & 58.46 & 60.09 & \textbf{60.83} & 56.80 \\
Citeseer     & 63.92 & 67.50 & \textbf{68.66} & 67.60 \\
CoCS         & 82.28 & 90.85 & \textbf{90.92} & 90.01 \\
CoPhysics    & 85.92 & 92.54 & 92.61 & \textbf{93.18} \\
Cora         & 59.95 & 77.82 & 53.58 & \textbf{79.13} \\
Cornell      & 69.73 & 66.49 & 68.65 & \textbf{78.38} \\
DBLP         & 67.34 & 70.13 & 66.42 & \textbf{73.12} \\
Deezer       & \textbf{55.22} & 52.13 & 52.31 & 54.73 \\
LastFMAsia   & 72.65 & 79.47 & 78.03 & \textbf{79.90} \\
Minesweeper  & 84.15 & 80.13 & 80.68 & \textbf{84.36} \\
Pubmed       & 75.12 & 77.46 & 74.98 & \textbf{87.17} \\
Questions    & 97.13 & 97.11 & 97.02 & \textbf{97.96} \\
RomanEmpire  & 69.80 & 64.06 & 66.36 & \textbf{75.68} \\
Texas        & \textbf{81.62} & 73.51 & 73.51 & 81.08 \\
Tolokers     & 78.22 & \textbf{78.24} & 78.12 & 78.16 \\
Wisconsin    & 73.33 & 71.77 & 61.18 & \textbf{84.31} \\
WikiCS       & 74.99 & 73.77 & 74.16 & \textbf{75.70} \\
\midrule
Average (\%) $\uparrow$      & 66.47 & 68.23 & 66.54 & \textbf{70.75} \\
Avg. Rank $\downarrow$ & 2.73 & 2.04 & 2.88 & \textbf{1.35} \\
\bottomrule
\end{tabular}
}
\label{tab:comparison_diverse}
\end{table*}

\section{Complete Results of Section 4.4 Across All Datasets}
\label{app:complete_results}

In this section, we provide the complete experimental settings and detailed per-dataset results for the cross-domain generalization evaluation in Section~\ref{sec:graphany}.

\paragraph{Dataset Categories.}
The evaluation covers 26 graph datasets spanning substantially different domains.
Specifically, the datasets are grouped into six representative categories:

\begin{itemize}[leftmargin=1.5em, topsep=2pt, parsep=0pt, itemsep=1pt]
\item Citation Networks: Cora, Citeseer, Pubmed, and ogbn-arxiv.
\item Co-authorship Networks: CoCS, CoPhysics, DBLP, and WikiCS.
\item Social Networks: Actor, BlogCatalog, Deezer, LastFMAsia, Questions, and Tolokers.
\item Co-purchase Graphs: AmzComp, AmzPhoto, and AmzRatings.
\item Heterophilic Benchmarks: Texas, Wisconsin, Cornell, RomanEmpire, Chameleon, and Minesweeper.
\item Air-traffic Graphs: AirBrazil, AirEU, and AirUS.
\end{itemize}

\paragraph{Baselines.}
We compare REEF with three representative fully-inductive graph learning baselines:
(1) GAT~\cite{gat}: A standard task-specific graph neural network baseline;
(2) GraphAny~\cite{zhao2024graphany}: A recent fully-inductive graph foundation model designed to generalize across arbitrary graphs with different feature and label spaces;
(3) TS-Mean~\cite{finkelshtein2025equivariance}:
A transfer-based graph foundation baseline using shared transferable representations.
For fair comparison, the results of GraphAny and TS-Mean are directly adopted from their original papers.

\paragraph{Evaluation Protocol.}
Unlike conventional transductive node classification, fully-inductive graph learning requires models to generalize across unseen graphs with substantially different graph structures, feature spaces, and semantic distributions.
This experiment is designed to evaluate whether REEF learns transferable relational knowledge rather than dataset-specific patterns.

As shown in Table~\ref{tab:comparison_diverse}, REEF achieves the best overall performance with an average accuracy of 70.75\%, outperforming GraphAny by 2.52\%, GAT by 4.28\%, and TS-Mean by 4.21\%. Moreover, REEF achieves the best average rank (1.35), indicating consistently strong performance across datasets rather than improvements concentrated on only a few benchmarks. More specifically, REEF demonstrates particularly strong improvements on challenging heterophily datasets such as RomanEmpire, Wisconsin, and Cornell, suggesting that relation-token modeling can better capture semantic interactions under weak homophily settings. In addition, REEF consistently performs well across substantially different domains, including citation networks, social networks, and co-purchase graphs, indicating that the learned relational representations are transferable beyond domain-specific feature statistics. Although some baselines achieve strong performance on specific datasets, REEF maintains more balanced performance across nearly all datasets, leading to the best overall average rank. Overall, these results demonstrate that REEF exhibits strong cross-domain transferability and robustness, supporting the effectiveness of relation-token-based graph foundation modeling.

\section{Experimental Details of LM-based Transfer Learning}
\label{app:lm_feature_transfer}

In this section, we provide detailed experimental settings for the LM-based transfer learning evaluation in Section~\ref{sec:lm_init}.

\paragraph{Datasets.}
The evaluation includes both citation-network and E-commerce text-attributed graph (TAG) datasets.

\begin{itemize}[leftmargin=1.5em, topsep=2pt, parsep=0pt, itemsep=1pt]

\item \textbf{Citation Networks:}
Nodes represent papers and edges denote citation relationships. Node features are derived from textual descriptions such as titles, abstracts, and keywords.

\begin{itemize}[leftmargin=1.5em]

\item Cora:
Contains 2,708 scientific publications from the computer science domain categorized into 7 classes.

\item CiteSeer:
Consists of 3,186 publications across 6 categories with textual metadata describing each paper.

\item Pubmed:
A biomedical citation network with 19,717 papers in 3 categories, where node text features are derived from diabetes-related abstracts.

\end{itemize}

\item \textbf{E-commerce Graphs:}
Nodes correspond to products with textual descriptions such as titles and categories, while edges represent semantic relations including co-purchase or rating interactions.

\begin{itemize}[leftmargin=1.5em]

\item History:
Contains 41,551 products and 503,180 co-purchase edges across 12 product categories.

\item Photo:
Comprises 48,362 products with 873,782 co-purchase edges and textual product descriptions.

\item Amazon Ratings:
A user--item bipartite graph containing 24,492 nodes and 186,100 rating edges spanning 5 product categories.

\end{itemize}

\end{itemize}

Comprehensive dataset statistics are available in the official TSGFM repository\footnote{\url{https://github.com/CurryTang/TSGFM}}.

\paragraph{Baselines.}
We compare REEF against representative text-space graph foundation models from TSGFM~\citep{tsgfm}, including:

\begin{itemize}[leftmargin=1.5em, topsep=2pt, parsep=0pt, itemsep=1pt]

\item \textbf{Graph Self-supervised Learning Methods:}
GraphMAE~\citep{hou2022graphmae}.

\item \textbf{Graph Prompt Learning Methods:}
GPPT~\citep{sun2022gppt},
Gprompt~\citep{liu2023graphprompt},
Prodigy~\citep{huang2024prodigy},
and OFA~\citep{liu2023one}.

\item \textbf{Graph Foundation Models:}
GIT~\cite{git}, which learns transferable graph generalities across tasks and domains via task-tree pretraining.

\item \textbf{LLM-based Graph Learning Methods:}
GraphLLM~\citep{graphllm},
GraphText~\citep{zhao2023graphtext},
GraphAdapter~\citep{graphadapter},
and LLaGA~\citep{chen2024llaga}.

\item \textbf{Lightweight Text Encoder Baseline:}
SBERT (no pretrain), which directly uses Sentence-BERT embeddings without graph pretraining.

\end{itemize}

\paragraph{Experimental Settings.}
Following the evaluation protocol of TSGFM~\citep{tsgfm}, we evaluate all methods under a 3-shot transfer learning setting, where only three labeled samples per class are available for downstream adaptation.
Node textual attributes are encoded into embeddings using Sentence-BERT~\citep{sentencebert}.

REEF is pretrained under two configurations:
(i) \textit{Setting~1}, using FB15K237, WN18RR, Pubmed, and Citeseer;
and
(ii) \textit{Setting~2}, which further includes the Photo dataset.
The second setting introduces E-commerce graph data during pretraining, enabling evaluation of domain-aware transferability.

Additional implementation details and hyperparameter settings follow the official TSGFM benchmark configuration.

\section{Broader Impacts}

The proposed framework
\model~has significant broader impacts for the development of graph foundation models (GFMs). 
By introducing relation tokens and adaptive hypernetworks, 
\model~offers a flexible and scalable approach to handle relational information across diverse datasets. This methodology not only improves the generalization and transferability of graph-based models but also provides insights into the potential of multi-dataset learning and graph data augmentation. The ability to adaptively generate parameters for aggregators, classifiers, and feature projectors across different domains paves the way for more robust and transferable GFM applications in fields such as knowledge graph construction, recommendation systems, and multi-modal learning. Additionally, the identification of scaling laws in \model~can guide future research on scaling GFMs for larger, more complex datasets, contributing to the broader field of AI-driven graph learning.

\section{Ethical Considerations}
This work focuses on methodological advances in graph foundation models.
All experiments are conducted on publicly available benchmark datasets and do not involve human subjects, private user data, or safety-critical decision-making systems.
We do not identify any direct ethical concerns associated with this work.

\section{The Use of Large Language Models (LLMs)}

In this paper, large language models were utilized exclusively for grammatical polishing and stylistic refinement, aimed at enhancing the clarity and readability of our presentation of results and conclusions.

\begin{table*}[hbtp]
\centering
\caption{Relations and Descriptions for Different Domains. ``Relation type'' of the knowledge graph is the corresponding dataset.}
\resizebox{\textwidth}{!}{
\begin{tabular}{|>{\centering\arraybackslash}m{2cm}|>{\centering\arraybackslash}m{3cm}|m{11cm}|}
\hline
\rowcolor[HTML]{EFEFEF} \textbf{Domain} & \textbf{Relation Type} & \textbf{Description} \\
\hline
\multirow{2}{*}[-0.5em]{\textbf{Citation}} 
& Citation & 
\textnormal{A citation in a network represents the act of one scientific paper referencing another, embodying the flow and accumulation of knowledge. It reflects the topical relevance between papers and researchers' acknowledgment of prior work, serving as a means to validate and support their own research while highlighting the collaborative nature of scientific inquiry.} \\
\cline{2-3}
& Paper classification & 
\textnormal{Paper classification in graph learning focuses on predicting the categories or topics of individual papers within a graph.} \\
\hline
\multirow{2}{*}[-1.3em]{\textbf{WebKB}} 
& Hyperlinks &
\textnormal{A hyperlink represents a relationship or connection between nodes (e.g., entities such as documents, web pages, or words). It encapsulates semantic associations, enables information flow, and reflects the nature of interactions, which may vary in type, intensity, or direction. }\\
\cline{2-3}
& Webpage classification & 
\textnormal{Webpage classification involves predicting the categories or types of individual webpages in a graph, often determined by the content, purpose, or role of the webpage within a network of hyperlinks.} \\
\hline
\multirow{2}{*}[-1.5em]{\textbf{Amazon}} 
& Co-purchase & 
\textnormal{The ``bought together'' relationship between goods represents the co-purchasing behavior of customers, indicating a semantic association between products based on shared purchasing patterns. This relationship reflects the likelihood of two products being bought in conjunction, capturing their complementary nature or relevance in a shopping context.} \\
\cline{2-3}
& Product classification & \textnormal{Product classification is a task in graph learning that predicts the categories or labels of product nodes, focusing on their classification based on attributes like type, function, or consumer category in a co-purchasing or recommendation graph.} \\
\hline
\multirow{2}{*}[-2.em]{
\textbf{\parbox[c][0.em][c]{1.7cm}{\centering Knowledge\\Graph}}} & 
(FB15K237) &
\textnormal{[
`/location/country/form\_of\_government', \newline
`/tv/tv\_program/regular\_cast./tv/regular\_tv\_appearance/actor', \newline 
`/media\_common/netflix\_genre/titles', \newline
`/award/award\_winner/awards\_won./award/award\_honor/award\_winner', \newline
`/soccer/football\_team/current\_roster./sports/sports\_team\_roster/position', \newline
`/soccer/football\_team/current\_roster./soccer/football\_roster\_position/position', \newline
`/film/actor/film./film/performance/film', \newline
... , \newline
] \newline
(Total 237 relations)
}
\\
\cline{2-3}
& (WN18RR) & 
\textnormal{[
`\_hypernym', `\_derivationally\_related\_form', `\_instance\_hypernym', \newline `\_also\_see', `\_member\_meronym', `\_synset\_domain\_topic\_of', `\_has\_part', `\_member\_of\_domain\_usage', `\_member\_of\_domain\_region', `\_verb\_group', 
`\_similar\_to'
] \newline
(Total 11 relations)
}
\\
\hline
\end{tabular}
}
\label{tab:relation_description}
\end{table*}


\newpage
\section*{NeurIPS Paper Checklist}

The checklist is designed to encourage best practices for responsible machine learning research, addressing issues of reproducibility, transparency, research ethics, and societal impact. Do not remove the checklist: {\bf The papers not including the checklist will be desk rejected.} The checklist should follow the references and follow the (optional) supplemental material.  The checklist does NOT count towards the page
limit. 

Please read the checklist guidelines carefully for information on how to answer these questions. For each question in the checklist:
\begin{itemize}
    \item You should answer \answerYes{}, \answerNo{}, or \answerNA{}.
    \item \answerNA{} means either that the question is Not Applicable for that particular paper or the relevant information is Not Available.
    \item Please provide a short (1--2 sentence) justification right after your answer (even for \answerNA). 
\end{itemize}

{\bf The checklist answers are an integral part of your paper submission.} They are visible to the reviewers, area chairs, senior area chairs, and ethics reviewers. You will also be asked to include it (after eventual revisions) with the final version of your paper, and its final version will be published with the paper.

The reviewers of your paper will be asked to use the checklist as one of the factors in their evaluation. While \answerYes{} is generally preferable to \answerNo{}, it is perfectly acceptable to answer \answerNo{} provided a proper justification is given (e.g., error bars are not reported because it would be too computationally expensive'' or ``we were unable to find the license for the dataset we used''). In general, answering \answerNo{} or \answerNA{} is not grounds for rejection. While the questions are phrased in a binary way, we acknowledge that the true answer is often more nuanced, so please just use your best judgment and write a justification to elaborate. All supporting evidence can appear either in the main paper or the supplemental material, provided in appendix. If you answer \answerYes{} to a question, in the justification please point to the section(s) where related material for the question can be found.

IMPORTANT, please:
\begin{itemize}
    \item {\bf Delete this instruction block, but keep the section heading ``NeurIPS Paper Checklist"},
    \item  {\bf Keep the checklist subsection headings, questions/answers and guidelines below.}
    \item {\bf Do not modify the questions and only use the provided macros for your answers}.
\end{itemize}


\begin{enumerate}

\item {\bf Claims}
    \item[] Question: Do the main claims made in the abstract and introduction accurately reflect the paper's contributions and scope?
    \item[] Answer: \answerYes{} 
    \item[] Justification:
    This paper introduces a relation-based graph foudation model, with the main contributions and scope outlined in the abstract and introduction.
    \item[] Guidelines:
    \begin{itemize}
        \item The answer \answerNA{} means that the abstract and introduction do not include the claims made in the paper.
        \item The abstract and/or introduction should clearly state the claims made, including the contributions made in the paper and important assumptions and limitations. A \answerNo{} or \answerNA{} answer to this question will not be perceived well by the reviewers. 
        \item The claims made should match theoretical and experimental results, and reflect how much the results can be expected to generalize to other settings. 
        \item It is fine to include aspirational goals as motivation as long as it is clear that these goals are not attained by the paper. 
    \end{itemize}

\item {\bf Limitations}
    \item[] Question: Does the paper discuss the limitations of the work performed by the authors?
    \item[] Answer: \answerYes{}{} 
    \item[] Justification:
     We analyze the limitations and future work in Section~\ref{sec:couclusion}.
    \item[] Guidelines:
    \begin{itemize}
        \item The answer \answerNA{} means that the paper has no limitation while the answer \answerNo{} means that the paper has limitations, but those are not discussed in the paper. 
        \item The authors are encouraged to create a separate ``Limitations'' section in their paper.
        \item The paper should point out any strong assumptions and how robust the results are to violations of these assumptions (e.g., independence assumptions, noiseless settings, model well-specification, asymptotic approximations only holding locally). The authors should reflect on how these assumptions might be violated in practice and what the implications would be.
        \item The authors should reflect on the scope of the claims made, e.g., if the approach was only tested on a few datasets or with a few runs. In general, empirical results often depend on implicit assumptions, which should be articulated.
        \item The authors should reflect on the factors that influence the performance of the approach. For example, a facial recognition algorithm may perform poorly when image resolution is low or images are taken in low lighting. Or a speech-to-text system might not be used reliably to provide closed captions for online lectures because it fails to handle technical jargon.
        \item The authors should discuss the computational efficiency of the proposed algorithms and how they scale with dataset size.
        \item If applicable, the authors should discuss possible limitations of their approach to address problems of privacy and fairness.
        \item While the authors might fear that complete honesty about limitations might be used by reviewers as grounds for rejection, a worse outcome might be that reviewers discover limitations that aren't acknowledged in the paper. The authors should use their best judgment and recognize that individual actions in favor of transparency play an important role in developing norms that preserve the integrity of the community. Reviewers will be specifically instructed to not penalize honesty concerning limitations.
    \end{itemize}

\item {\bf Theory assumptions and proofs}
    \item[] Question: For each theoretical result, does the paper provide the full set of assumptions and a complete (and correct) proof?
    \item[] Answer: \answerNA{} 
    \item[] Justification: 
    This paper does not include theoretical results.
    \item[] Guidelines:
    \begin{itemize}
        \item The answer \answerNA{} means that the paper does not include theoretical results. 
        \item All the theorems, formulas, and proofs in the paper should be numbered and cross-referenced.
        \item All assumptions should be clearly stated or referenced in the statement of any theorems.
        \item The proofs can either appear in the main paper or the supplemental material, but if they appear in the supplemental material, the authors are encouraged to provide a short proof sketch to provide intuition. 
        \item Inversely, any informal proof provided in the core of the paper should be complemented by formal proofs provided in appendix or supplemental material.
        \item Theorems and Lemmas that the proof relies upon should be properly referenced. 
    \end{itemize}

    \item {\bf Experimental result reproducibility}
    \item[] Question: Does the paper fully disclose all the information needed to reproduce the main experimental results of the paper to the extent that it affects the main claims and/or conclusions of the paper (regardless of whether the code and data are provided or not)?
    \item[] Answer: \answerYes{} 
    \item[] Justification: 
     We provide a link to the anonymized code in the abstract, along with the hyperparameters used in Appendix D. The results in our paper are reproducible.
    \item[] Guidelines:
    \begin{itemize}
        \item The answer \answerNA{} means that the paper does not include experiments.
        \item If the paper includes experiments, a \answerNo{} answer to this question will not be perceived well by the reviewers: Making the paper reproducible is important, regardless of whether the code and data are provided or not.
        \item If the contribution is a dataset and\slash or model, the authors should describe the steps taken to make their results reproducible or verifiable. 
        \item Depending on the contribution, reproducibility can be accomplished in various ways. For example, if the contribution is a novel architecture, describing the architecture fully might suffice, or if the contribution is a specific model and empirical evaluation, it may be necessary to either make it possible for others to replicate the model with the same dataset, or provide access to the model. In general. releasing code and data is often one good way to accomplish this, but reproducibility can also be provided via detailed instructions for how to replicate the results, access to a hosted model (e.g., in the case of a large language model), releasing of a model checkpoint, or other means that are appropriate to the research performed.
        \item While NeurIPS does not require releasing code, the conference does require all submissions to provide some reasonable avenue for reproducibility, which may depend on the nature of the contribution. For example
        \begin{enumerate}
            \item If the contribution is primarily a new algorithm, the paper should make it clear how to reproduce that algorithm.
            \item If the contribution is primarily a new model architecture, the paper should describe the architecture clearly and fully.
            \item If the contribution is a new model (e.g., a large language model), then there should either be a way to access this model for reproducing the results or a way to reproduce the model (e.g., with an open-source dataset or instructions for how to construct the dataset).
            \item We recognize that reproducibility may be tricky in some cases, in which case authors are welcome to describe the particular way they provide for reproducibility. In the case of closed-source models, it may be that access to the model is limited in some way (e.g., to registered users), but it should be possible for other researchers to have some path to reproducing or verifying the results.
        \end{enumerate}
    \end{itemize}

\item {\bf Open access to data and code}
    \item[] Question: Does the paper provide open access to the data and code, with sufficient instructions to faithfully reproduce the main experimental results, as described in supplemental material?
    \item[] Answer: \answerYes{} 
    \item[] Justification:
    We provide our code, as well as links to public datasets.
    \item[] Guidelines:
    \begin{itemize}
        \item The answer \answerNA{} means that paper does not include experiments requiring code.
        \item Please see the NeurIPS code and data submission guidelines (\url{https://neurips.cc/public/guides/CodeSubmissionPolicy}) for more details.
        \item While we encourage the release of code and data, we understand that this might not be possible, so \answerNo{} is an acceptable answer. Papers cannot be rejected simply for not including code, unless this is central to the contribution (e.g., for a new open-source benchmark).
        \item The instructions should contain the exact command and environment needed to run to reproduce the results. See the NeurIPS code and data submission guidelines (\url{https://neurips.cc/public/guides/CodeSubmissionPolicy}) for more details.
        \item The authors should provide instructions on data access and preparation, including how to access the raw data, preprocessed data, intermediate data, and generated data, etc.
        \item The authors should provide scripts to reproduce all experimental results for the new proposed method and baselines. If only a subset of experiments are reproducible, they should state which ones are omitted from the script and why.
        \item At submission time, to preserve anonymity, the authors should release anonymized versions (if applicable).
        \item Providing as much information as possible in supplemental material (appended to the paper) is recommended, but including URLs to data and code is permitted.
    \end{itemize}

\item {\bf Experimental setting/details}
    \item[] Question: Does the paper specify all the training and test details (e.g., data splits, hyperparameters, how they were chosen, type of optimizer) necessary to understand the results?
    \item[] Answer: \answerYes{} 
    \item[] Justification: 
    We provide experimental details in Appendix D.
    \item[] Guidelines:
    \begin{itemize}
        \item The answer \answerNA{} means that the paper does not include experiments.
        \item The experimental setting should be presented in the core of the paper to a level of detail that is necessary to appreciate the results and make sense of them.
        \item The full details can be provided either with the code, in appendix, or as supplemental material.
    \end{itemize}

\item {\bf Experiment statistical significance}
    \item[] Question: Does the paper report error bars suitably and correctly defined or other appropriate information about the statistical significance of the experiments?
    \item[] Answer: \answerYes{} 
    \item[] Justification: We report the standard deviation of the results.
    \item[] Guidelines:
    \begin{itemize}
        \item The answer \answerNA{} means that the paper does not include experiments.
        \item The authors should answer \answerYes{} if the results are accompanied by error bars, confidence intervals, or statistical significance tests, at least for the experiments that support the main claims of the paper.
        \item The factors of variability that the error bars are capturing should be clearly stated (for example, train/test split, initialization, random drawing of some parameter, or overall run with given experimental conditions).
        \item The method for calculating the error bars should be explained (closed form formula, call to a library function, bootstrap, etc.)
        \item The assumptions made should be given (e.g., Normally distributed errors).
        \item It should be clear whether the error bar is the standard deviation or the standard error of the mean.
        \item It is OK to report 1-sigma error bars, but one should state it. The authors should preferably report a 2-sigma error bar than state that they have a 96\% CI, if the hypothesis of Normality of errors is not verified.
        \item For asymmetric distributions, the authors should be careful not to show in tables or figures symmetric error bars that would yield results that are out of range (e.g., negative error rates).
        \item If error bars are reported in tables or plots, the authors should explain in the text how they were calculated and reference the corresponding figures or tables in the text.
    \end{itemize}

\item {\bf Experiments compute resources}
    \item[] Question: For each experiment, does the paper provide sufficient information on the computer resources (type of compute workers, memory, time of execution) needed to reproduce the experiments?
    \item[] Answer: \answerYes{} 
    \item[] Justification: This paper provide information on the computer resources in Appendix D.
    \item[] Guidelines:
    \begin{itemize}
        \item The answer \answerNA{} means that the paper does not include experiments.
        \item The paper should indicate the type of compute workers CPU or GPU, internal cluster, or cloud provider, including relevant memory and storage.
        \item The paper should provide the amount of compute required for each of the individual experimental runs as well as estimate the total compute. 
        \item The paper should disclose whether the full research project required more compute than the experiments reported in the paper (e.g., preliminary or failed experiments that didn't make it into the paper). 
    \end{itemize}
    
\item {\bf Code of ethics}
    \item[] Question: Does the research conducted in the paper conform, in every respect, with the NeurIPS Code of Ethics \url{https://neurips.cc/public/EthicsGuidelines}?
    \item[] Answer: \answerYes{} 
    \item[] Justification:
    The research conducted in the paper conform, in every respect, with the NeurIPS Code of Ethics.
    \item[] Guidelines:
    \begin{itemize}
        \item The answer \answerNA{} means that the authors have not reviewed the NeurIPS Code of Ethics.
        \item If the authors answer \answerNo, they should explain the special circumstances that require a deviation from the Code of Ethics.
        \item The authors should make sure to preserve anonymity (e.g., if there is a special consideration due to laws or regulations in their jurisdiction).
    \end{itemize}

\item {\bf Broader impacts}
    \item[] Question: Does the paper discuss both potential positive societal impacts and negative societal impacts of the work performed?
    \item[] Answer: \answerYes{} 
    \item[] Justification: 
    We discuss the broader impacts of this paper in Appendix F.
    \item[] Guidelines:
    \begin{itemize}
        \item The answer \answerNA{} means that there is no societal impact of the work performed.
        \item If the authors answer \answerNA{} or \answerNo, they should explain why their work has no societal impact or why the paper does not address societal impact.
        \item Examples of negative societal impacts include potential malicious or unintended uses (e.g., disinformation, generating fake profiles, surveillance), fairness considerations (e.g., deployment of technologies that could make decisions that unfairly impact specific groups), privacy considerations, and security considerations.
        \item The conference expects that many papers will be foundational research and not tied to particular applications, let alone deployments. However, if there is a direct path to any negative applications, the authors should point it out. For example, it is legitimate to point out that an improvement in the quality of generative models could be used to generate Deepfakes for disinformation. On the other hand, it is not needed to point out that a generic algorithm for optimizing neural networks could enable people to train models that generate Deepfakes faster.
        \item The authors should consider possible harms that could arise when the technology is being used as intended and functioning correctly, harms that could arise when the technology is being used as intended but gives incorrect results, and harms following from (intentional or unintentional) misuse of the technology.
        \item If there are negative societal impacts, the authors could also discuss possible mitigation strategies (e.g., gated release of models, providing defenses in addition to attacks, mechanisms for monitoring misuse, mechanisms to monitor how a system learns from feedback over time, improving the efficiency and accessibility of ML).
    \end{itemize}
    
\item {\bf Safeguards}
    \item[] Question: Does the paper describe safeguards that have been put in place for responsible release of data or models that have a high risk for misuse (e.g., pre-trained language models, image generators, or scraped datasets)?
    \item[] Answer: \answerNA{} 
    \item[] Justification: 
    Our paper has no such risks.
    \item[] Guidelines:
    \begin{itemize}
        \item The answer \answerNA{} means that the paper poses no such risks.
        \item Released models that have a high risk for misuse or dual-use should be released with necessary safeguards to allow for controlled use of the model, for example by requiring that users adhere to usage guidelines or restrictions to access the model or implementing safety filters. 
        \item Datasets that have been scraped from the Internet could pose safety risks. The authors should describe how they avoided releasing unsafe images.
        \item We recognize that providing effective safeguards is challenging, and many papers do not require this, but we encourage authors to take this into account and make a best faith effort.
    \end{itemize}

\item {\bf Licenses for existing assets}
    \item[] Question: Are the creators or original owners of assets (e.g., code, data, models), used in the paper, properly credited and are the license and terms of use explicitly mentioned and properly respected?
    \item[] Answer: \answerYes{} 
    \item[] Justification: 
     The paper properly credits the creators or original owners of assets and
respects the license and terms of use.
    \item[] Guidelines:
    \begin{itemize}
        \item The answer \answerNA{} means that the paper does not use existing assets.
        \item The authors should cite the original paper that produced the code package or dataset.
        \item The authors should state which version of the asset is used and, if possible, include a URL.
        \item The name of the license (e.g., CC-BY 4.0) should be included for each asset.
        \item For scraped data from a particular source (e.g., website), the copyright and terms of service of that source should be provided.
        \item If assets are released, the license, copyright information, and terms of use in the package should be provided. For popular datasets, \url{paperswithcode.com/datasets} has curated licenses for some datasets. Their licensing guide can help determine the license of a dataset.
        \item For existing datasets that are re-packaged, both the original license and the license of the derived asset (if it has changed) should be provided.
        \item If this information is not available online, the authors are encouraged to reach out to the asset's creators.
    \end{itemize}

\item {\bf New assets}
    \item[] Question: Are new assets introduced in the paper well documented and is the documentation provided alongside the assets?
    \item[] Answer: \answerYes{} 
    \item[] Justification: New assets introduced in this paper are well documented.
    \item[] Guidelines:
    \begin{itemize}
        \item The answer \answerNA{} means that the paper does not release new assets.
        \item Researchers should communicate the details of the dataset\slash code\slash model as part of their submissions via structured templates. This includes details about training, license, limitations, etc. 
        \item The paper should discuss whether and how consent was obtained from people whose asset is used.
        \item At submission time, remember to anonymize your assets (if applicable). You can either create an anonymized URL or include an anonymized zip file.
    \end{itemize}

\item {\bf Crowdsourcing and research with human subjects}
    \item[] Question: For crowdsourcing experiments and research with human subjects, does the paper include the full text of instructions given to participants and screenshots, if applicable, as well as details about compensation (if any)? 
    \item[] Answer: \answerNA{} 
    \item[] Justification:     The paper does not involve crowdsourcing nor research with human subjects.
    \item[] Guidelines:
    \begin{itemize}
        \item The answer \answerNA{} means that the paper does not involve crowdsourcing nor research with human subjects.
        \item Including this information in the supplemental material is fine, but if the main contribution of the paper involves human subjects, then as much detail as possible should be included in the main paper. 
        \item According to the NeurIPS Code of Ethics, workers involved in data collection, curation, or other labor should be paid at least the minimum wage in the country of the data collector. 
    \end{itemize}

\item {\bf Institutional review board (IRB) approvals or equivalent for research with human subjects}
    \item[] Question: Does the paper describe potential risks incurred by study participants, whether such risks were disclosed to the subjects, and whether Institutional Review Board (IRB) approvals (or an equivalent approval/review based on the requirements of your country or institution) were obtained?
    \item[] Answer: \answerNA{} 
    \item[] Justification: The paper does not involve crowdsourcing nor research with human subjects.
    \item[] Guidelines:
    \begin{itemize}
        \item The answer \answerNA{} means that the paper does not involve crowdsourcing nor research with human subjects.
        \item Depending on the country in which research is conducted, IRB approval (or equivalent) may be required for any human subjects research. If you obtained IRB approval, you should clearly state this in the paper. 
        \item We recognize that the procedures for this may vary significantly between institutions and locations, and we expect authors to adhere to the NeurIPS Code of Ethics and the guidelines for their institution. 
        \item For initial submissions, do not include any information that would break anonymity (if applicable), such as the institution conducting the review.
    \end{itemize}

\item {\bf Declaration of LLM usage}
    \item[] Question: Does the paper describe the usage of LLMs if it is an important, original, or non-standard component of the core methods in this research? Note that if the LLM is used only for writing, editing, or formatting purposes and does \emph{not} impact the core methodology, scientific rigor, or originality of the research, declaration is not required.
    \item[] Answer: \answerNA{} 
    \item[] Justification:  We only use LLMs to improve writing.
    \item[] Guidelines:
    \begin{itemize}
        \item The answer \answerNA{} means that the core method development in this research does not involve LLMs as any important, original, or non-standard components.
        \item Please refer to our LLM policy in the NeurIPS handbook for what should or should not be described.
    \end{itemize}

\end{enumerate}

\end{document}